\documentclass[letterpaper]{article}
\usepackage[preprint]{aaai2027}
\usepackage[hyphens]{url}
\usepackage{graphicx}
\usepackage{natbib}
\usepackage{caption}
\usepackage{booktabs}
\usepackage{amsmath}
\title{Wnuan: Staged Post-Training for Question Answering over Proprietary Enterprise Knowledge}
\author{
Xiaofeng Shi\textsuperscript{1}\equalcontrib\thanks{\raggedright Corresponding author: Xiaofeng Shi, xfshi@baai.ac.cn.},\quad Xiaosong Qiu\textsuperscript{1}\equalcontrib,\quad Wenxin Ma\textsuperscript{2},\quad Qian Kou\textsuperscript{1},\\
Yiming Pan\textsuperscript{3}\thanks{\raggedright Work completed during an internship at Beijing Academy of Artificial Intelligence (BAAI).},\quad Longbin Yu\textsuperscript{1},\quad Ying Liu\textsuperscript{1},\quad Haiping Wang\textsuperscript{1},\quad Hua Zhou\textsuperscript{1}\thanks{\raggedright Project leader.}
}
\affiliations{
\textsuperscript{1}Beijing Academy of Artificial Intelligence (BAAI)\\
\textsuperscript{2}Beijing District Heating Group Co., Ltd. (BDHG)\\
\textsuperscript{3}Beijing University of Posts and Telecommunications (BUPT)
}

\begin{document}

\maketitle

\begin{abstract}
Enterprise question answering requires models to acquire proprietary knowledge without discarding general capabilities. We present Wnuan, a three-stage pipeline that constructs task-oriented supervision from documents, performs supervised fine-tuning with general-data replay, and applies reinforcement learning to residual errors. On the 707-question WnuanBench, the primary 32B route raises acceptable-answer rate (AAR) from 52.76\% before adaptation to 80.06\% after SFT and 91.51\% after RL. Under a matched 100-update protocol, residual-error sampling outperforms full-pool and size-matched random sampling by 3.11 and 2.97 points, respectively. Source-cluster bootstrap intervals remain above zero for both contrasts, and a same-domain validation set preserves the ordering. The general-benchmark average decreases by 5.17 points across the route, concentrated in instruction following. The automatic evaluation ensemble agrees with an authoritative domain expert on 90.5\% of a stratified Wnuan-Inst response sample. These results characterize both the gains and the general-capability cost of staged enterprise adaptation.
\end{abstract}

\section{Introduction}

Enterprise question answering depends on internal policies, technical standards, and operating procedures that are often absent from public pretraining data. Adapting a general-purpose language model to this setting requires the model to learn proprietary knowledge, retain general instruction-following ability, and use a limited post-training budget efficiently.

Prior work addresses these requirements separately. Task-oriented corpus adaptation converts documents into learnable supervision \citep{cheng2024adapting}. Post-training may change generalization and instruction following \citep{kirk2024rlhf,lin2024alignmenttax}, while mixing pretraining-data updates into RLHF has reduced public-benchmark regressions \citep{ouyang2022training}. Retrieval-augmented generation (RAG) supplies evidence at inference time \citep{lewis2020rag,zhang2024raft}. Data-selection methods choose influential examples before instruction tuning or filter uninformative groups during RL \citep{xia2024less,yu2025dapo}. Less is known about how these choices interact in a single enterprise QA pipeline, especially after SFT has already corrected most easy examples.

\begin{figure*}[t]
    \centering
    \includegraphics[width=\textwidth]{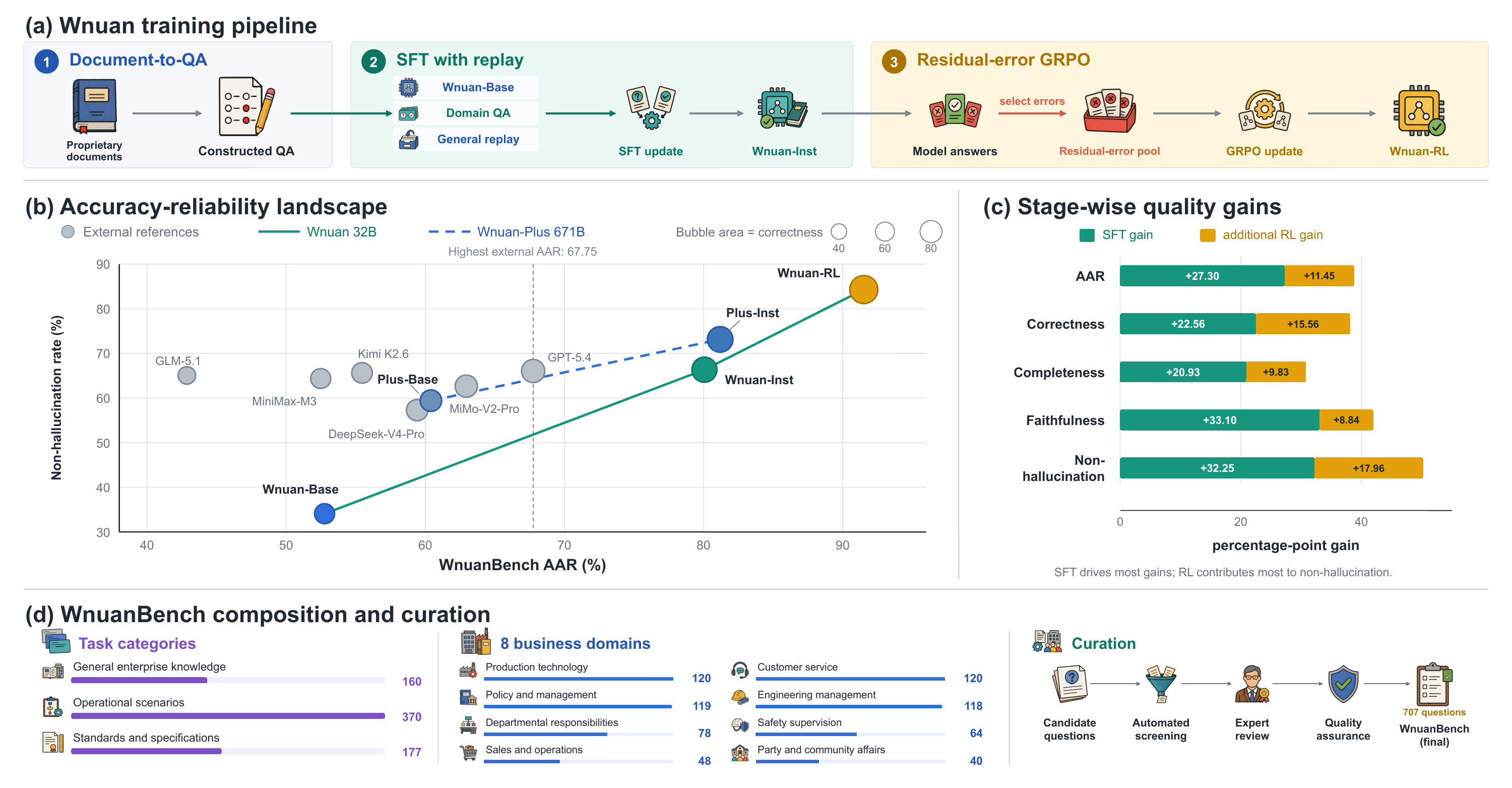}
    \caption{Wnuan training path, primary 32B results, contextual systems, and WnuanBench construction. (a) Enterprise documents are converted into task-oriented QA supervision, used for SFT with general-data replay, and then revisited through residual-error RL. (b) Horizontal position is WnuanBench AAR, vertical position is non-hallucination rate, and bubble area encodes correctness. Gray API points and the dashed 671B connector provide context. The solid 32B route is the primary trajectory. (c) Stacked bars decompose the Base-to-RL endpoint differences into Base-to-Inst and Inst-to-RL increments. (d) WnuanBench follows a benchmark-specific screening, expert-review, and quality-assurance path distinct from training-QA construction.}
    \label{fig:overview}
\end{figure*}

We train Wnuan in three stages (Figure~\ref{fig:overview}). Stage I converts enterprise documents into self-contained question--answer supervision and rewrites eligible answers in a form aligned with the target model. Stage II performs full-parameter SFT with general-data replay. Stage III identifies examples that Wnuan-Inst still answers incorrectly and applies semantic-reward GRPO to those residual errors. We evaluate retrieval separately rather than training a retrieval-aware generator.

The paper centers on the complete enterprise-model training pipeline and the resulting WnuanBench evaluation. Stage-wise studies measure the contribution of SFT, the domain--general trade-off induced by replay, and the gains and instruction-following cost of residual-error RL. A fixed-budget experiment compares residual-error, full-pool, and size-matched random sampling. Public general benchmarks, a same-domain validation set, and the training-side validation signal support development.

We contribute an end-to-end post-training pipeline that converts proprietary documents into a closed-book enterprise QA model, selects a general-data replay operating point, and applies residual-error RL. We also introduce WnuanBench and use it to evaluate the primary 32B training trajectory under an automatic correctness ensemble calibrated on 147 Wnuan-Inst responses labeled by one domain expert. Configuration studies, a controlled three-arm GRPO experiment, source-cluster sensitivity analysis, and general-capability measurements identify where the pipeline gains accuracy and where it loses instruction-following performance.

\section{Related Work}

\paragraph{Domain adaptation and task-oriented supervision.}
Domain QA synthesis spans AdaptLLM's reading-comprehension reformulation, pre-instruction tuning, and knowledge- or coverage-aware generation in KEFT and DS2-Instruct \citep{cheng2024adapting,jiang2024knowledge,li2025keft,xu2026ds2}. Wnuan assembles these precedents as a recipe, not a new synthesis method.

\paragraph{Retention during specialization.}
Specialization can change generalization and instruction following \citep{kirk2024rlhf,lin2024alignmenttax}. Replay is a documented mitigation in RLHF, and domain-knowledge injection work likewise mixes general QA examples during fine-tuning \citep{ouyang2022training,bhushan2025systematic}. We measure an SFT replay grid and select an observed domain--general operating point rather than propose a new retention objective.

\paragraph{RL and data selection.}
PPO and GRPO provide the optimization basis for modern language-model post-training \citep{schulman2017ppo,ouyang2022training,shao2024deepseekmath,deepseek2025r1}. LESS selects influential instruction examples, while DAPO filters zero-advantage prompt groups online \citep{xia2024less,yu2025dapo}. Difficulty-aware alignment shows that examples can exceed model capacity, whereas fixed-budget GRPO studies also report benefits from prioritizing hard prompts \citep{gao2025difficulty,pikus2025hard}. Wnuan makes a simpler recipe choice: it selects examples still judged incorrect after SFT and compares that offline pool with full-pool and size-matched random sampling in enterprise QA. It does not propose a general data-selection algorithm.

\paragraph{Retrieval and model-based evaluation.}
RAG augments generation with non-parametric memory \citep{lewis2020rag}. RAFT trains models to use relevant evidence while ignoring distractors \citep{zhang2024raft}. Wnuan is not retrieval-aware, so we compare its checkpoints with a fixed retrieval-concatenation baseline. Because open-form enterprise answers cannot be scored reliably by exact match, we use a multi-model judging procedure and calibrate its final binary decision against a domain expert, following the broader literature on LLM-based evaluation \citep{zheng2023judging,liu2023geval,zhu2025judgelm}.

\section{Method}

\subsection{Problem Setting and Metric}

Let $\mathcal{D}$ be a collection of proprietary enterprise documents and let $\mathcal{T}=\{(x_i,y_i)\}$ be a QA pool derived from those documents. The goal is to produce a closed-book instruction model that answers questions from the represented knowledge base while retaining useful general behavior.

Our primary outcome is the \emph{acceptable-answer rate} (AAR):
\begin{equation}
\mathrm{AAR}=\frac{N_{\mathrm{correct}}+N_{\mathrm{partially\ correct}}}{N}.
\label{eq:aar}
\end{equation}
Equation~\ref{eq:aar} merges full and partial credit. AAR is not a strict fully-correct rate. We use AAR throughout the paper even though the stored evaluation field is named \texttt{Accuracy}.
Here, $N$ is the number of evaluated questions, while $N_{\mathrm{correct}}$ and $N_{\mathrm{partially\ correct}}$ count the questions assigned the corresponding final ensemble labels.

\subsection{Stage I: Document-to-QA Data Construction}

The available pre-rewriting QA inventory contains 231,662 rows, 221,825 unique questions, 5,648 source paths, and 38,467 source-chunk identifiers. The reference implementation first segments OCR-normalized documents at semantic and paragraph boundaries. It then extracts a named anchor, generates a self-contained question from one of six task forms, produces an answer from the supporting chunk, and filters candidates using rule, referent, answerability, faithfulness, and quality checks.

Eligible answers are subsequently rewritten by the target model. A rewritten answer replaces the original only when the two answers pass a semantic-similarity gate and the candidate passes format filtering. The resulting SFT domain set contains 221,294 examples. We call this operation \emph{target-aligned answer rewriting}. The Stage-I experiments do not show an independent domain-AAR gain from rewriting. The construction thresholds, model roles, retained counts, and provenance boundary are detailed in Supplementary Appendix C. The historical files do not preserve row-level generator lineage.

Among the final SFT examples, 164,793 candidate generations pass the similarity gate. Format filtering removes 49 candidates containing the literal token \texttt{\string<think\string>}, leaving 164,744 rewritten targets and 56,550 retained original targets.

\subsection{Stage II: SFT with General-Data Replay}

The main 32B route starts from Qwen3-32B, which we denote Wnuan-Base \citep{yang2025qwen3}. Wnuan-Inst is trained on the 221,294 domain examples, 106,950 public general examples \citep{soren2025chineseqwen3}, and smaller auxiliary instruction, train-out, and identity sets. We measure nominal replay levels of 0\%, 5\%, 25\%, 50\%, and 100\%. These are display labels relative to the number of domain examples. The selected 50\% setting contains 106,950 general examples, or an actual ratio of 48.3\%.

We select the replay setting with the highest unweighted average of MMLU, IFEval, and C-Eval in the measured grid \citep{hendrycks2021mmlu,zhou2023ifeval,huang2023ceval}. The same-domain validation result is a secondary development check. This rule selects the nominal 50\% setting and defines Wnuan-Inst, the common initialization for Stage III. The complete replay grid and its component benchmark scores are provided in Supplementary Appendix E.

\subsection{Stage III: Residual-Error RL}

Residual selection uses a 230,183-row QA pool with stored original targets. Let $\mathcal{P}$ denote this selection pool, $f_{\mathrm{inst}}$ denote Wnuan-Inst, and $J_{\mathrm{sel}}$ denote the recorded residual-selection judge. We construct
\begin{equation}
\mathcal{P}_{\mathrm{err}}=\{(x_i,y_i)\in\mathcal{P}:J_{\mathrm{sel}}(f_{\mathrm{inst}}(x_i),y_i)=\mathrm{incorrect}\},
\label{eq:residual-pool}
\end{equation}
Equation~\ref{eq:residual-pool} selects 56,147 examples.

The main RL run applies GRPO with five rollouts per prompt. Its semantic reward is
\begin{equation}
\begin{aligned}
r &= 0.6r_{\mathrm{acc}}+0.3r_{\mathrm{quality}}+0.1r_{\mathrm{format}},\\
r_{\mathrm{quality}}&=(r_{\mathrm{logic}}+r_{\mathrm{prof}}+r_{\mathrm{concise}})/3.
\end{aligned}
\label{eq:reward}
\end{equation}
We adapt the semantic reward in Equation~\ref{eq:reward} from MechVQA to text-only enterprise QA \citep{kou2026mechvqa}. Each component is normalized to $[0,1]$. The terms score answer correctness ($r_{\mathrm{acc}}$), logical soundness ($r_{\mathrm{logic}}$), professional expression ($r_{\mathrm{prof}}$), concision ($r_{\mathrm{concise}}$), and compliance with the required answer tags ($r_{\mathrm{format}}$). A locally deployed Qwen3.5-35B judge \citep{qwen2026qwen35} scores the semantic components, while normalized exact matches take a deterministic unit-score fast path. For each five-response group, GRPO standardizes rewards within the group and optimizes a token-level PPO-style objective with clipping $\epsilon=0.2$, dual-clip coefficient $C=3$, and reference-policy penalty $\beta=10^{-2}$. The complete objective, reward definitions, and run configurations are specified in Supplementary Appendix D.

We compare data selection in a separate direct-answer experiment. The residual-error, full-pool, and size-matched random arms share the Wnuan-Inst initialization, prompt, scoring procedure, rollout count, optimizer settings, and 100-update schedule. The direct-answer prompt omits the tags checked by the format scorer, so $r_{\mathrm{format}}=0$ for every arm. The comparison uses the common $0.6r_{\mathrm{acc}}+0.3r_{\mathrm{quality}}$ signal. Each arm follows two 50-update segments, with model weights retained and the optimizer restarted at the midpoint. The complete schedule and controlled-arm endpoint analysis appear in Supplementary Appendix F.

\section{Experimental Setup}

\subsection{Development Validation, WnuanBench, and Evaluation}

Development uses a validation set sampled from the same enterprise-domain distribution as the training data. WnuanBench contains 707 questions grounded in formal enterprise documents: 160 general-knowledge, 370 operational-scenario, and 177 standards/specification questions across eight business domains. Internal personnel curate its questions and references independently of the automated training-QA pipeline. Each record includes a question, reference answer, source, and domain label. No QA record from the validation set or WnuanBench enters training or residual selection. The validation set supports domain-side development, whereas WnuanBench is reserved for final evaluation after the recipe and endpoint are frozen. Because both sets draw on the represented enterprise knowledge base, this is an in-domain evaluation rather than a test of source-held-out or cross-enterprise generalization.

For formal evaluation, two primary judges, gpt-oss-120b \citep{openai2025gptoss} and MiniMax-M2.5 \citep{minimax2026m25}, assign ordered correctness labels on $\{0,0.5,1\}$. DeepSeek-V3.2 \citep{deepseekai2025v32} supplies a third vote on disagreement, and the ordered median is retained. One domain expert labels a stratified sample of Wnuan-Inst responses. On the 147 valid labels, the final binary decision agrees with the expert on 90.5\% of responses (95\% CI: 85.7--94.6\%; $\kappa=0.796$); the post-stratified estimate is 90.4\%. The three-class decision matches exactly on 103 responses. Of the remaining decisions, 31 automatic labels are more generous and 13 are stricter. Only two disagreements cross directly between correct and incorrect. The evaluation rubric, adjudication procedure, and expert-calibration study are described in Supplementary Appendices A--B.

The expert was selected for authority over the governing documents and access to the relevant enterprise context. Hallucination detection has precision 0.868, recall 0.657, and F1 0.748 on the same sample. The calibration supports the binary AAR decision more directly than the auxiliary labels.

Primary paired confidence intervals use 2,000 question-level bootstrap resamples. A sensitivity analysis additionally resamples the 217 source documents with replacement while retaining all questions from each sampled source. Paired binary comparisons use McNemar tests, with Holm adjustment for the three planned residual/full/random contrasts. All controlled Stage-III configurations use one fixed training protocol, and the random arm uses a size-matched subset drawn with seed 42. The estimand is the paired difference between the completed endpoints under that protocol.

\subsection{Training and Comparison Conditions}

The final SFT run uses full-parameter bf16 training with DeepSpeed ZeRO-3 on 32 accelerators, a global batch of 32, a peak learning rate of $10^{-5}$, a 1,024-token cutoff with packing, three epochs, and 11,976 optimizer updates. The main GRPO run uses 2 nodes $\times$ 8 H100-80GB accelerators, global batch 128, learning rate $10^{-6}$, and three epochs.

Each controlled data-selection arm starts from Wnuan-Inst and follows the shared schedule described above. The full arm samples from all 230,183 selection-pool rows. The residual and random arms each contain 56,147 rows. Residual versus random controls pool size. Residual versus full instead tests sampling efficiency under a common update budget, not equal per-example exposure. These three arms form the matched comparison in the paper. The API systems in Table~\ref{tab:main-results}, identified by their official releases \citep{zai2026glm51,minimax2026m3,moonshotai2026kimik26,deepseekai2026v4,xiaomi2026mimov2pro,openai2026gpt54}, and the 671B LoRA-SFT route provide context only. The GRPO configurations and controlled-arm diagnostics appear in Supplementary Appendices D and F, while Appendices I and J document the retrieval analysis and 671B route.

\begin{table*}[t]
\centering
\small
\begin{tabular}{llrrrrrr}
\toprule
Checkpoint & Backbone / adaptation & AAR & Correct. & Complete. & Faithful. & Halluc. $\downarrow$ & General avg. \\
\midrule
\multicolumn{8}{l}{\textit{External API references (descriptive)}} \\
GLM-5.1 & API reference & 42.86 & 30.76 & 29.28 & 39.04 & 34.94 & -- \\
MiniMax-M3 & API reference & 52.48 & 40.03 & 36.78 & 39.04 & 35.64 & -- \\
Kimi K2.6 & API reference & 55.45 & 42.64 & 40.38 & 44.55 & 34.37 & -- \\
DeepSeek-V4-Pro & API reference & 59.41 & 46.11 & 42.36 & 42.50 & 42.72 & -- \\
MiMo-V2-Pro & API reference & 62.94 & 49.15 & 44.63 & 45.90 & 37.34 & -- \\
GPT-5.4 & API reference & 67.75 & 53.61 & 44.77 & 52.05 & 33.95 & -- \\
\midrule
\multicolumn{8}{l}{\textit{Wnuan checkpoints}} \\
Wnuan-Base & Qwen3-32B & 52.76 & 40.31 & 36.00 & 30.20 & 65.91 & 88.61 \\
Wnuan-Inst & 32B, full SFT & 80.06 & 62.87 & 56.93 & 63.30 & 33.66 & 84.64 \\
Wnuan-RL & 32B, SFT + GRPO & \textbf{91.51} & \textbf{78.43} & \textbf{66.76} & \textbf{72.14} & \textbf{15.70} & 83.44 \\
\midrule
Wnuan-Plus-Base & DeepSeek-V3.1-Terminus & 60.40 & 46.53 & 45.54 & 44.63 & 40.59 & \textbf{90.38} \\
Wnuan-Plus-Inst & 671B, LoRA-SFT & 81.19 & 65.28 & 55.73 & 69.45 & 26.87 & 84.34 \\
\bottomrule
\end{tabular}
\caption{Closed-book WnuanBench results (\%). AAR is the primary outcome. General avg. is the unweighted mean of MMLU, IFEval, and C-Eval, whose components are listed in Supplementary Appendix G. API systems and the 671B LoRA-SFT route provide contextual endpoints because their decoding, compute, backbone, and adaptation conditions are not matched to the primary 32B route.}
\label{tab:main-results}
\end{table*}

\begin{figure*}[t]
    \centering
    \includegraphics[width=0.80\textwidth]{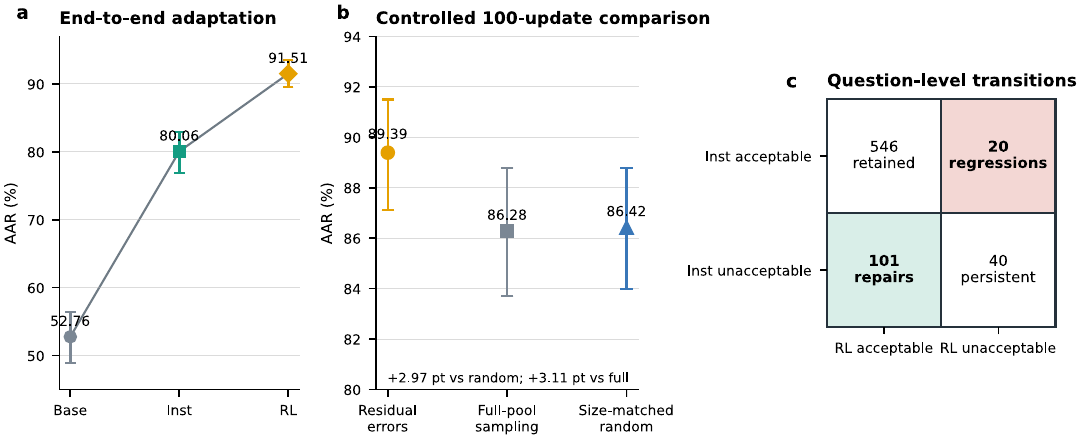}
    \caption{Core evidence. (a) Closed-book AAR rises from Wnuan-Base to Wnuan-Inst to Wnuan-RL. Intervals resample benchmark questions. (b) Under a common 100-update protocol, residual-error sampling outperforms full-pool and size-matched random sampling. (c) From Wnuan-Inst to Wnuan-RL, 101 initially unacceptable answers become acceptable and 20 initially acceptable answers regress.}
    \label{fig:core-evidence}
\end{figure*}

\section{Results}

\subsection{Enterprise Leaderboard and End-to-End Adaptation}

Table~\ref{tab:main-results} provides the WnuanBench leaderboard, while Figures~\ref{fig:core-evidence}(a) and~\ref{fig:overview}(b--c) summarize the primary 32B training trajectory. Wnuan-Base reaches 52.76\% AAR (95\% CI: 48.9--56.4). SFT raises AAR to 80.06\% (76.9--82.9), a paired gain of 27.30 points (95\% CI: 23.20--31.54; $p<0.001$). RL raises it further to 91.51\% (89.5--93.5), a gain of 11.45 points over Wnuan-Inst (95\% CI: 8.49--14.43; $p<0.001$). Source-cluster bootstrap intervals are 19.71--34.21 points for the SFT gain and 8.52--15.05 for the RL gain. On the separate validation set, the same Inst-to-RL transition increases AAR from 76.89\% to 89.00\% ($+12.11$ points), closely matching the final WnuanBench gain. Across the ensemble endpoints, completeness increases from 36.00\% to 66.76\%, faithfulness from 30.20\% to 72.14\%, and hallucination decreases from 65.91\% to 15.70\%.

The final column summarizes general retention. The general average changes from 88.61\% for Wnuan-Base to 84.64\% for Wnuan-Inst and 83.44\% for Wnuan-RL. The component trajectories in Supplementary Appendix G show that SFT decreases all three scores, whereas from Wnuan-Inst to Wnuan-RL, MMLU increases by 0.61 points and C-Eval by 2.29 points while IFEval decreases by 6.52 points. The domain gains accompany a concentrated instruction-following cost rather than a uniform decline.

\subsection{Residual-Error Sampling}

Figure~\ref{fig:core-evidence}(b) reports the controlled experiment under this common direct-answer GRPO protocol. Residual-error sampling reaches 89.39\% AAR, compared with 86.28\% when prompts are sampled from the full pool and 86.42\% for a size-matched random subset. Relative to random sampling, residual-error sampling gains 2.97 points (95\% question-level CI: 0.85--5.09; Holm-adjusted $p=0.035$). Relative to full-pool sampling, it gains 3.11 points (0.71--5.66; adjusted $p=0.035$). The corresponding source-cluster intervals are 0.81--5.08 and 0.83--5.70 points. Full-pool and random sampling are statistically indistinguishable ($-0.14$ points; adjusted $p=1$).

The source-cluster analysis resamples all questions associated with each sampled source document and preserves both controlled residual contrasts above zero. Residual selection uses stored original targets, whereas SFT may use rewritten targets. A fixed-prediction audit remaps targets without new model answers: 91.69\% of examples whose mapped rewritten target differs from the original retain the same automated incorrect/acceptable membership. The incorrect rate changes by only $-0.86$ points, although 4,928 examples leave and 4,001 enter the residual set. This supports stability of the aggregate policy, not exact row-level invariance. The full migration table and sensitivity protocol appear in Supplementary Appendix F.

We also evaluate the same frozen endpoints on the same-domain validation set. The ordering is unchanged: residual-error, size-matched random, and full-pool sampling obtain 81.33\%, 78.67\%, and 77.78\% AAR. Residual versus full gains 3.56 points (95\% CI: 1.33--5.78; Holm-adjusted $p=0.008$). Residual versus random gains 2.67 points (0.22--5.00), with Holm-adjusted $p=0.068$ after correcting the three contrasts.

The residual-versus-random contrast controls pool size because both pools contain 56,147 examples. The residual-versus-full contrast evaluates whether concentrating a fixed update budget on current errors is more effective than drawing from the complete pool. Aggregate logs over updates 51--100 show lower on-policy accuracy reward but larger mean absolute PPO-KL and gradient-norm statistics for the residual arm, a pattern consistent with harder sampled prompts. The endpoint contrasts support residual-error sampling as the Stage-III data policy. The update-level statistics describe the associated training dynamics.

Both question sets, training-signal diagnostics, endpoint intervals, source-cluster sensitivity, and auxiliary answer-quality dimensions are reported in Supplementary Appendix F.

\begin{figure*}[t]
    \centering
    \includegraphics[width=0.76\textwidth]{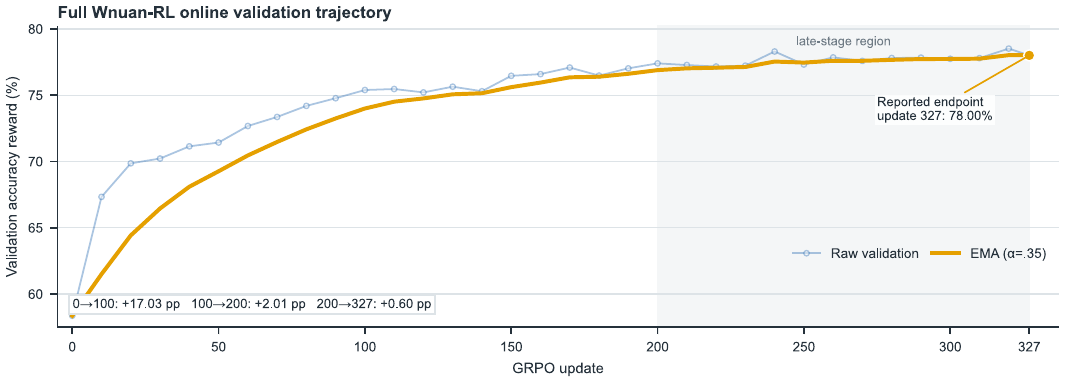}
    \caption{Online validation accuracy-reward trajectory for the main Wnuan-RL run. After the preliminary 100-update experiment selected residual-error sampling, the full run gains 17.03 points through update 100, 2.01 more through update 200, and 0.60 more through update 327. The reported endpoint lies in the shaded late-stage region.}
    \label{fig:main-grpo-trajectory}
\end{figure*}

\subsection{Training Dynamics and Stopping}

Figure~\ref{fig:main-grpo-trajectory} shows that the main run's validation accuracy reward rises from 58.36\% at initialization to 75.39\% at update 100, 77.40\% at update 200, and 78.00\% at update 327. The diminishing increments and compute budget determine the practical stop. Across the four controlled arms, the mean absolute gap between the update-100 validation reward and final WnuanBench correctness is 1.19 points, and validation gains preserve the observed ordering of final AAR gains. A comparison of the four-arm monitoring trajectories with their completed endpoints appears in Supplementary Appendix F; the correlations are descriptive rather than estimates over retraining variability.

The main run's overall validation reward rises from 0.5952 to 0.8332. Decomposition of that $0.2380$ increase attributes $0.1179$ to accuracy, $0.0415$ to mean semantic quality, and $0.0787$ to format compliance after applying the reward weights. Format reward reaches 1.0, so the aggregate reward gain is not interchangeable with correctness. This decomposition applies to the tagged-answer main run; the direct-answer controlled arms have zero format reward and compare data selection under their shared accuracy-and-quality signal. The reward components and distinct response formats used by the two experiments are reported in Supplementary Appendix D.

\subsection{Stage-I and Stage-II Design Studies}

\begin{figure*}[t]
    \centering
    \includegraphics[width=0.76\textwidth]{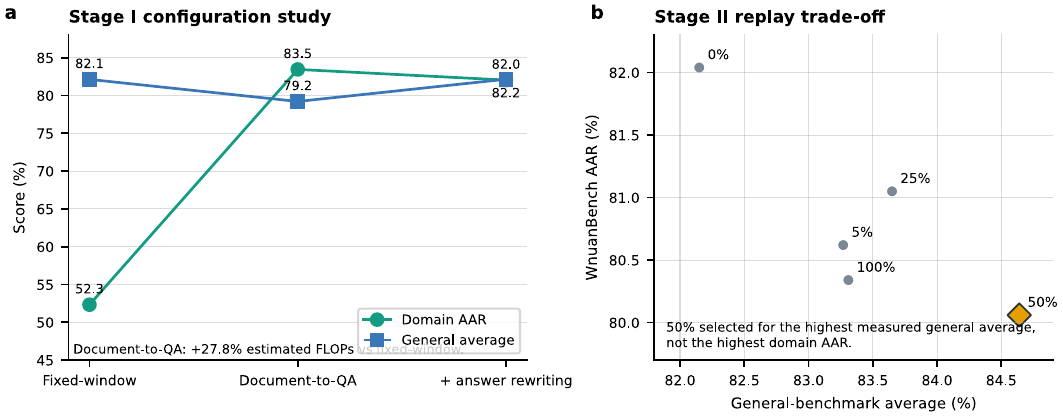}
    \caption{Design studies preceding Stage III. (a) Task-oriented Document-to-QA training substantially outperforms fixed-window training while using 27.8\% more estimated FLOPs. Target-aligned answer rewriting approximately preserves domain AAR while recovering part of the general-benchmark loss. (b) The replay grid exposes a domain--general trade-off, with the selected setting determined by the measured general-benchmark average.}
    \label{fig:stage-designs}
\end{figure*}

Figure~\ref{fig:stage-designs}(a) shows that Document-to-QA training improves AAR from 52.33\% to 83.45\% relative to fixed-window training, a 31.12-point gain obtained with 27.8\% more estimated FLOPs. Target-aligned answer rewriting retains 82.04\% AAR while restoring the general-benchmark average from 79.21\% to 82.15\%. These unequal-budget runs provide configuration-level evidence rather than an isolated causal estimate of QA organization. The complete outcome, budget, runtime, and uncertainty breakdown appears in Supplementary Appendix E.

Figure~\ref{fig:stage-designs}(b) summarizes the Stage-II replay trade-off. Relative to no replay, the selected 48.3\% operating point sacrifices 1.98 AAR points while gaining 2.49 points on the general-benchmark average. It was selected for the highest measured general average, not the highest domain AAR. The complete grid and component benchmarks appear in Supplementary Appendix E.

\subsection{What Stage III Changes}

Figure~\ref{fig:core-evidence}(c) gives a question-level view: Wnuan-RL repairs 101 Wnuan-Inst errors and regresses on 20 previously acceptable answers, for a net reduction of 81 errors. The remaining failures often involve exact numbers, dates, responsible departments, document names, closed lists, and omitted conditions. These categories are qualitative because WnuanBench does not contain mutually exclusive expert error labels. The full transition accounting appears in Supplementary Appendix G, and descriptive within-enterprise domain slices appear in Appendix H.

All eight business-domain slices improve numerically from Wnuan-Inst to Wnuan-RL, but only engineering management and departmental responsibilities remain significant after Holm correction. No controlled residual-versus-full or residual-versus-random domain contrast is significant. The domain analysis is descriptive, and the aggregate paired comparison remains primary. Domain sample sizes, adjusted tests, and controlled contrasts are reported in Supplementary Appendix H.

\subsection{Train-Free Retrieval Diagnostic}

For train-free context, we apply one fixed BM25+BGE-M3 top-5 retrieval-concatenation pipeline to all three checkpoints \citep{robertson2009bm25,chen2024bgem3}. RAG changes AAR from 52.76\% to 72.98\% for Wnuan-Base, from 80.06\% to 76.24\% for Wnuan-Inst, and from 91.51\% to 81.75\% for Wnuan-RL. Thus retrieval helps the unadapted model but is non-additive after SFT and RL under this pipeline. On the 433-question proxy slice with at least one same-domain retrieved chunk, retrieval improves Base and Inst but slightly reduces RL; on the remaining 274 questions, it reduces all three, most sharply after specialization. Domain match is not a gold relevance label, but the separation motivates confidence gating rather than unconditional concatenation. Because the diagnostic uses one retriever and historical no-RAG generations, it does not establish a general training--retrieval interaction. The paired RAG results, supporting quality dimensions, and retrieval-trace proxy slices are reported in Supplementary Appendix I.

\section{Discussion}

The Wnuan pipeline assigns a distinct operational role to each stage. Document-to-QA supervision organizes enterprise knowledge for closed-book learning, general-data replay selects a retention operating point, and residual-error GRPO concentrates the final update budget on remaining mistakes. The evidence has a corresponding hierarchy: Stage I is an unequal-compute configuration study, Stage II is a finite operating-point search, and Stage III contains the matched data-selection experiment. The paper therefore supports an end-to-end recipe and a controlled claim about residual-error sampling, not a compute-matched additive decomposition of all three stages.

The negative results also matter. RL reduces domain errors but lowers IFEval. A second regression-aware continuation does not recover that loss: relative to Wnuan-RL, AAR changes from 91.51\% to 91.37\%, hallucination rises from 15.70\% to 20.93\%, and IFEval falls from 80.00\% to 76.00\%. Without an otherwise identical unbucketed control, this experiment does not isolate the regression-aware partition rule. It nevertheless shows that another residual-focused continuation is not automatically beneficial and favors explicit instruction replay or a revised factual-consistency reward. The RL-2 configuration and endpoint comparison appear in Supplementary Appendix K.

Retrieval remains an inference-time intervention whose value depends on the checkpoint and retrieved context. It should be gated independently of the training recipe rather than treated as an automatically additive fourth stage.

\section{Limitations and Responsible Use}

\paragraph{Evaluation scope.}
The study covers one enterprise and in-domain validation and benchmark sets. WnuanBench is QA-record-disjoint from training and residual selection but shares the authorized source corpus; no source-, time-, enterprise-, or open-world split is available. Its benchmark-specific curation path is separate from automated training-QA generation. Question-level bootstrap intervals quantify uncertainty within this fixed benchmark, not over new documents or organizations.

\paragraph{Training evidence.}
Stage I is an unequal-budget configuration study, whereas the residual/full/random arms share a common 100-update protocol. The 91.51\% main endpoint and 89.39\% controlled residual endpoint differ in response format, batch size, sequence length, hardware, and schedule. Attribution is restricted to the three completed sampling arms; uncertainty-, loss-, influence-, and online zero-advantage selectors were not tested. Source-cluster intervals provide a sensitivity analysis, and evaluation calibration relies on Wnuan-Inst responses labeled by one domain expert.

\paragraph{Retrieval and reproducibility.}
The RAG diagnostic uses one retrieval pipeline, no gold Recall@5 labels, and unmatched generation seeds. Private documents and the complete benchmark cannot be released. The evaluation flow is documented in Supplementary Appendix L. The Code and Data Supplement provides the correctness prompt, a synthetic fixture, and reference statistics. Data construction and training used locally deployed models inside the controlled environment.

The intended use is internal knowledge assistance with human verification, not automated personnel, compliance, safety, or other high-impact decisions.

\paragraph{AI assistance disclosure.}
Generative AI tools supported language editing and consistency checks. The authors verified the text, references, figures, and conclusions and take responsibility for the submitted material.

\section{Conclusion}

Wnuan combines document-to-QA supervision, general-data replay, and residual-error RL for closed-book enterprise QA. On WnuanBench, the primary 32B route raises AAR from 52.76\% to 80.06\% after SFT and 91.51\% after RL. Under the matched 100-update protocol, residual-error sampling outperforms full-pool and size-matched random sampling, with both source-cluster intervals above zero. These results indicate that, after SFT resolves many easy examples, concentrating a fixed update budget on remaining errors is an effective policy under the tested protocol.

The gains come with clear boundaries. General-data replay trades some domain accuracy for broader capability retention, while RL reduces IFEval performance. WnuanBench measures mastery of the represented enterprise knowledge base rather than source-held-out or cross-enterprise transfer. Future work should improve instruction-following retention and test the pipeline under transfer-oriented and retrieval-aware settings.

\bibliography{references}
\clearpage
\appendix
\section*{Appendix Overview}
The appendices provide the implementation details and analyses supporting the main paper. Appendices A--B describe WnuanBench and judge calibration. Appendices C--F cover data construction, training, and stage-wise experiments. Appendices G--I analyze capability retention, business domains, and retrieval. Appendices J--L record the Wnuan-Plus configuration, the unsuccessful RL-2 extension, reproducibility, and responsible use.

The primary metric is the \emph{acceptable-answer rate} (AAR), the fraction labeled correct or partially correct. Evaluation exports name this field \texttt{Accuracy}. The paper uses AAR to distinguish it from strict full correctness.

\section{Development Validation, WnuanBench, and Statistical Protocol}

\subsection{Benchmark Composition}

WnuanBench contains 707 questions grounded in authorized enterprise documents. The task partition contains 160 general-knowledge, 370 operational-scenario, and 177 standards/specification questions. A separate taxonomy assigns the same questions to eight business domains. Each record includes a question, reference answer, source identifier, and domain label.

\subsection{Construction and Data Roles}

Internal personnel constructed WnuanBench independently of the automated training-QA pipeline. Candidate questions were checked against authorized enterprise documents for relevance, determinate answers, self-contained wording, and operational usefulness before expert review and quality assurance. Questions, answers, sources, domains, and available evidence excerpts were locked before final comparison. The fitting pipeline removes exact matches against the validation and benchmark QA snapshots before Stage-I/II training and Stage-III residual selection. Shared sources, enterprise facts, and semantically related questions remain because WnuanBench measures mastery of the represented knowledge base, not unseen-document transfer.

A validation set sampled from the training-domain distribution supports offline development. Public MMLU, IFEval, and C-Eval scores select the Stage-II replay setting, and the training-side validation signal supports Stage-III monitoring. WnuanBench is QA-record-disjoint from fitting and development data and is reserved for final evaluation. Table~\ref{tab:data-role-audit-v3} lists these roles. Exact matching finds no shared question between the validation set and WnuanBench.

The overlap analysis uses a portable pre-filter candidate-pool snapshot. Normalized exact matching removes 1,479 rows covering 698 of the 707 WnuanBench questions. After removal, BGE-M3 nearest-neighbor cosine similarity has a median of 0.9295, and 458 benchmark questions have a nearest retained training question at or above 0.90. This semantic proximity is consistent with the in-corpus evaluation setting.

\begin{table*}[t]
\centering
\small
\begin{tabular}{lrlccl}
\toprule
Artifact & Questions & Fitting & Dev. / selection & Final & SHA-256 prefix / identity \\
\midrule
Stage-specific fitting pools & Varies & Yes & No & No & Evaluation QA excluded pre-fitting \\
Same-domain validation & 900 & No & Yes & No & \texttt{f33f650f1f37} \\
WnuanBench & 707 & No & No & Yes & \texttt{ceeb79b0cc02} \\
MMLU / IFEval / C-Eval & Published sets & No & Stage-II replay & No & Released versions \\
\bottomrule
\end{tabular}
\caption{Data-role audit. Counts and digest prefixes identify the two frozen private evaluation snapshots. The validation and WnuanBench snapshots have zero exact-question overlap.}
\label{tab:data-role-audit-v3}
\end{table*}

On the validation set, Wnuan-Inst and Wnuan-RL obtain 76.89\% and 89.00\% AAR, respectively. Table~\ref{tab:validation-transition-v3} shows that the 12.11-point gain is accompanied by improvements in every supporting dimension and is close to the 11.45-point WnuanBench gain. AAR improves on 131 questions and regresses on 22 (exact McNemar $p=4.42\times10^{-20}$). All nine validation domains have non-negative AAR changes.

\begin{table*}[t]
\centering
\small
\begin{tabular}{lrrrr}
\toprule
Metric & Wnuan-Inst & Wnuan-RL & Difference & 95\% CI \\
\midrule
AAR & 76.89 & 89.00 & +12.11 & +9.56 to +14.67 \\
Correctness & 57.33 & 73.17 & +15.83 & +13.56 to +18.17 \\
Completeness & 51.78 & 60.94 & +9.17 & +7.06 to +11.39 \\
Faithfulness & 56.22 & 73.50 & +17.28 & +15.00 to +19.78 \\
Hallucination $\downarrow$ & 37.89 & 17.00 & $-20.89$ & $-24.11$ to $-17.78$ \\
\bottomrule
\end{tabular}
\caption{Paired Wnuan-Inst-to-Wnuan-RL changes on the separate validation set (\%). Intervals use the same question-level bootstrap protocol as the final benchmark.}
\label{tab:validation-transition-v3}
\end{table*}

\subsection{Evaluation Dimensions}

\paragraph{Acceptable-answer rate.}
AAR maps correct and partially correct responses to acceptable and incorrect responses to unacceptable. It is the primary outcome.

\paragraph{Supporting dimensions.}
Correctness is the mean ordered score on $\{0,0.5,1\}$. Completeness measures key-point coverage. Faithfulness measures support from the stored evidence field. Hallucination is the fraction of responses containing unsupported facts and is lower-is-better. The general benchmarks are MMLU, IFEval, and C-Eval. Their unweighted mean is used only for replay selection and summary analysis.

Mean ordered correctness on $\{0,0.5,1\}$ is reported alongside AAR for every principal endpoint, so the supporting score retains the distinction between full and partial credit even though AAR is the primary operational decision rate.

\subsection{Aggregation and Statistical Tests}

Raw votes and the supporting dimensions are stored, but the main inference uses AAR.

Primary paired confidence intervals use 2,000 question-level bootstrap resamples with seed 20260708. The source-cluster sensitivity analysis uses 10,000 resamples with the same seed, samples 217 source documents with replacement, and retains every question attached to each sampled source. McNemar tests use continuity correction when there are at least 25 discordant pairs and the exact binomial test otherwise. The RAG/no-RAG tests are exact. Holm adjustment is applied within planned comparison families.

\begin{figure*}[!t]
\centering
\includegraphics[width=\textwidth]{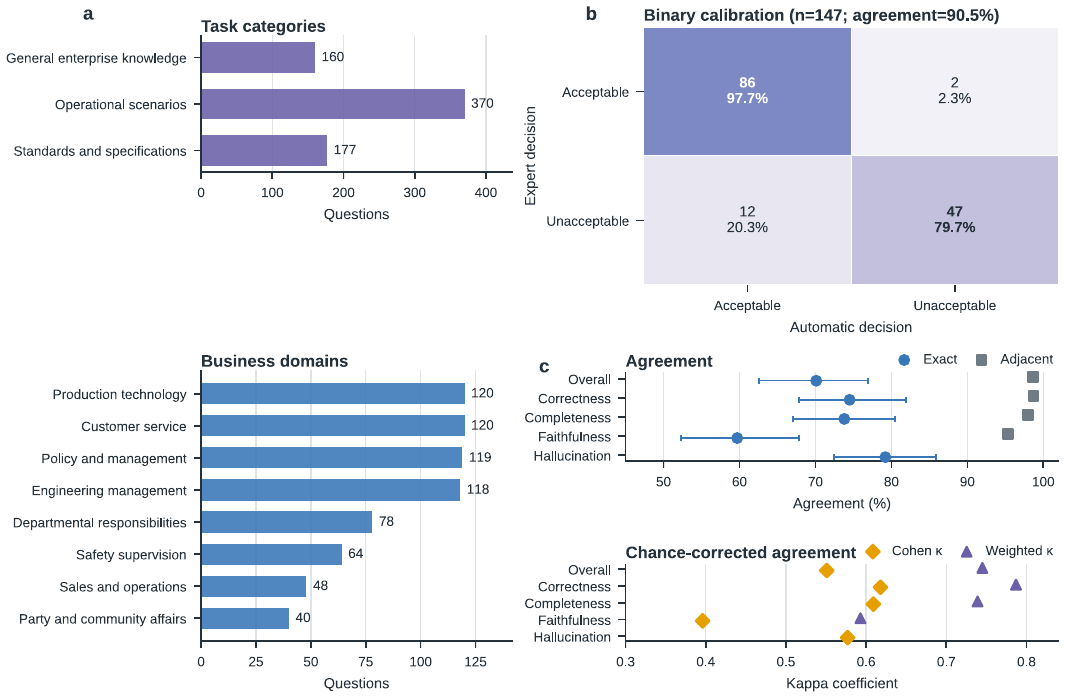}
\caption{Benchmark coverage and judge calibration. (a) Task-category and business-domain counts. (b) Binary confusion matrix between the automatic ensemble and one authoritative domain expert on 147 valid labels from the stratified evaluation sample. (c) Agreement percentages and kappa coefficients are shown in separate facets.}
\label{fig:supp-benchmark-judging}
\end{figure*}

Figure~\ref{fig:supp-benchmark-judging} summarizes the benchmark partitions and the scope of the available judge calibration.

\section{Calibration of Automatic Judging}

\subsection{Sampling and Binary Agreement}

The calibration sample contains 50 automatically correct, 50 automatically partial, and 50 automatically incorrect Wnuan-Inst responses from the formal evaluation results. One independent authoritative domain expert assigned final labels. Three missing overall labels leave $n=147$. Because the sample is balanced by the automatic label, unweighted agreement is the primary summary. Post-stratification is reported as a sensitivity check.

The expert was selected for domain authority, access to restricted enterprise context, and responsibility for interpreting the governing documents. The expert's final labels provide the human reference for calibration.

The automatic acceptable/unacceptable decision agrees with the expert on 133 of 147 responses, or 90.5\% (95\% CI: 85.7--94.6\%). Cohen's $\kappa$ is 0.796. The post-stratified agreement estimate is 90.4\%. These values quantify agreement with the authoritative expert labels.

\subsection{Ordinal and Auxiliary Agreement}

For the three-class overall label, 103 decisions match exactly, 31 automatic decisions are more generous, and 13 are stricter. Only two disagreements cross directly between correct and incorrect. Most occur at the correct/partial or partial/incorrect boundaries. Hallucination detection has precision 0.868, recall 0.657, and F1 0.748 against the expert's positive labels.

\subsection{Selection, Training, and Evaluation Judge Roles}

The residual-selection judge $J_{\mathrm{sel}}$ and formal evaluation use the same adaptive three-model correctness protocol. Locally deployed gpt-oss-120b and MiniMax-M2.5 provide primary labels on $\{0,0.5,1\}$ \citep{openai2025gptoss,minimax2026m25}; on disagreement, DeepSeek-V3.2 supplies a third vote and the ordered median of valid scores is retained \citep{deepseekai2025v32}. Stage III selects aggregate 0, corresponding to incorrect. A separate locally deployed Qwen3.5-35B judge produces GRPO semantic rewards \citep{qwen2026qwen35}. The reward judge is therefore disjoint, whereas selection and formal evaluation share the full ensemble.

\section{Stage I: Data Construction and Target-Aligned Answer Rewriting}

\subsection{Data Inventory and Provenance}

The pre-rewriting file contains 231,662 rows, 221,825 unique questions, 5,648 source paths, and 38,467 source-chunk identifiers. The final SFT-domain file contains 221,294 rows, 220,915 unique questions, 5,536 source paths, and 38,359 source-chunk identifiers. The 10,368-row reduction reflects consolidation of repeated question instances and exact-match filtering.

The final rows do not retain anchor, task-form, generator-version, or complete run-lineage fields. Tables~\ref{tab:data-construction-config-v3} and~\ref{tab:data-model-roles-v3} report the configuration and functional assignments available from the reference implementation.

\subsection{Reference Implementation}

\begin{table}[t]
\centering
\small
\begin{tabular}{p{0.56\columnwidth}p{0.30\columnwidth}}
\toprule
Configuration item & Value \\
\midrule
Minimum file length & 50 characters \\
Semantic chunk length & 1,000--4,000 characters \\
Adjacent-chunk overlap & 400 characters \\
File quality threshold & 2.0 / 5 \\
Chunk quality threshold & 2.0 / 5 \\
Minimum question score & 3.0 / 5 \\
Minimum answer score & 3.0 / 5 \\
Answer-rewriting gate & cosine $\geq 0.8$ \\
\bottomrule
\end{tabular}
\caption{Recorded configuration of the reference data-construction implementation. The composite QA score assigns quality buckets and is not an additional $0.6$ discard threshold.}
\label{tab:data-construction-config-v3}
\end{table}

File- and chunk-quality scores are produced by an external scorer. The reference code is fail-open when that endpoint errors or returns no score. The 2.0 thresholds apply to successful responses and do not prove that every stored row received a valid external score. Endpoint-failure counts, stage-by-stage retention, exact endpoint versions, configuration hashes, and row-level provenance were not retained.

\begin{table*}[t]
\centering
\small
\begin{tabular}{p{0.27\textwidth}p{0.24\textwidth}p{0.40\textwidth}}
\toprule
Pipeline role & Model or rule & Recorded behavior \\
\midrule
File and chunk screening & \raggedright IndustryCorpus2 DataRater \citep{xiaofeng_shi_2026} & Regression score with the thresholds in Table~\ref{tab:data-construction-config-v3} \\
Semantic chunking & Qwen3-14B & Semantic and paragraph-boundary segmentation \\
Question generation and validation & Qwen3-32B & Anchor-aware generation followed by self-containedness checks \\
Answer generation & DeepSeek-V3.2 & Single-model generation; optional voting disabled \\
Referent check and QA evaluation & Qwen3-32B & Rule, referent, answerability, faithfulness, and quality checks \\
\bottomrule
\end{tabular}
\caption{Functional assignments in the reference implementation. They document the available code, not row-level lineage for the final training file.}
\label{tab:data-model-roles-v3}
\end{table*}

The generator supports six task forms: fact extraction, mechanism explanation, design rationale, conditional constraint, limitation or trade-off, and comparison. Questions must name the anchor explicitly, avoid local references such as ``the above,'' and remain answerable without the source document. Candidate pairs are removed when rule or referent checks fail, the question is judged unanswerable, the answer is judged unfaithful, or question/answer quality falls below 3 on a five-point scale.

\subsection{Answer Rewriting}

The target model receives the question and original answer and generates a candidate answer. The candidate replaces the original only when MiniLM cosine similarity is at least 0.8 and format filtering passes. Generation uses Qwen3-32B with temperature 0.7, top-$p$ 0.8, top-$k$ 20, and a 4,096-token output limit.

Of 221,294 rows, 164,793 candidate generations (74.47\%) pass the similarity gate. Format filtering removes 49 otherwise eligible candidates containing the literal token \texttt{\string<think\string>}. The final artifact contains 164,744 rewritten targets (74.45\%) and 56,550 retained original targets. We use \emph{target-aligned answer rewriting} rather than \emph{knowledge induction} for this operation because the controlled evidence does not show an independent domain-AAR gain.

\section{Training, GRPO, and Retrieval Configurations}

\subsection{Supervised Fine-Tuning}

\begin{table*}[t]
\centering
\small
\begin{tabular}{p{0.20\textwidth}p{0.72\textwidth}}
\toprule
Component & Configuration \\
\midrule
Initialization & Qwen3-32B; full-parameter SFT in bf16 \\
Domain data & 221,294 Document-to-QA examples after answer rewriting and exact-question exclusion \\
General replay & 106,950 examples from Chinese-Qwen3-235B-Thinking-2507-Distill-100k \\
Other data & 9,747 instruction + 2,387 train-out + 1,042 identity examples; 341,420 examples in total \\
Sequence construction & 1,024-token cutoff; packing enabled; 127,722 packed sequences \\
Optimization & AdamW; weight decay 0; max gradient norm 1; 3 epochs; 11,976 updates; global batch 32; peak LR $10^{-5}$; cosine decay; warmup ratio 0.05 \\
Randomness & Training seed 42 \\
Parallelism & DeepSpeed ZeRO-3; 4 nodes $\times$ 8 accelerators \\
Software & Transformers 4.53.0; PyTorch 2.6.0+cu124; Datasets 3.6.0; Tokenizers 0.21.4 \\
\bottomrule
\end{tabular}
\caption{Configuration of the final Wnuan-Inst run. The SFT log does not record the accelerator model.}
\label{tab:sft-config-v3}
\end{table*}

Table~\ref{tab:sft-config-v3} records the Wnuan-Inst configuration. The general-replay data are the complete 106,950-example train split of \path{Jackrong/Chinese-Qwen3-235B-Thinking-2507-Distill-100k}, released under Apache-2.0.

\subsection{Main and Controlled GRPO Runs}

Both RL experiments initialize from Wnuan-Inst and sample five responses per prompt. Table~\ref{tab:rl-config-v3} separates the reported main run from the controlled data-selection experiment.

\begin{table*}[t]
\centering
\small
\begin{tabular}{lll}
\toprule
Item & Main Wnuan-RL & Controlled data-selection experiment \\
\midrule
Prompt / response limit & 2,048 / 4,096 tokens & 2,048 / 2,048 tokens \\
Response form & Reasoning plus tagged answer & Direct answer; reasoning disabled \\
Training pool & 56,147 Wnuan-Inst errors & Error, full, or size-matched random pool \\
Update budget & 3 epochs; reported update 327 & 100 updates for every arm \\
Global / rollout batch & 128 / 512 & 80 / 480 \\
Rollouts per prompt & 5 & 5 \\
Learning rate / temperature & $10^{-6}$ / 1.0 & $10^{-6}$ / 1.0 \\
Clip ratio / KL coefficient & 0.2 / $10^{-2}$ & 0.2 / $10^{-2}$ \\
Tensor parallelism & 4 & 4 \\
Hardware allocation & 2 nodes $\times$ 8 H100-80GB & 5 nodes $\times$ 8 A100-SXM4-40GB \\
\bottomrule
\end{tabular}
\caption{Configurations of the main GRPO run and the controlled data-selection experiment.}
\label{tab:rl-config-v3}
\end{table*}

For each prompt $x$, GRPO samples responses $y_i\sim\pi_{\theta_{\mathrm{old}}}(\cdot\mid x)$ and sets $R_i=r(x,y_i)$. With five responses per group, let $\bar R$ and $s_R$ denote the mean and standard deviation of the five rewards. The normalized advantage is
\begin{equation}
\widehat A_i=\frac{R_i-\bar R}{s_R+10^{-6}},\qquad
\bar R=\frac{1}{5}\sum_{j=1}^{5}R_j.
\label{eq:supp-advantage-v3}
\end{equation}
Let $\rho_{i,t}=\pi_\theta(y_{i,t}\mid x,y_{i,<t})/\pi_{\theta_{\mathrm{old}}}(y_{i,t}\mid x,y_{i,<t})$. The implementation maximizes
\begin{equation}
\mathcal{J}(\theta)=
\frac{\sum_{i,t}m_{i,t}\left[\psi(\rho_{i,t},\widehat A_i)-\beta d^{\mathrm{LV}}_{i,t}\right]}
{\sum_{i,t}m_{i,t}},
\label{eq:supp-grpo-objective-v3}
\end{equation}
In Equation~\ref{eq:supp-grpo-objective-v3}, $m_{i,t}$ masks valid response tokens and
\begin{equation}
\psi(\rho,A)=
\begin{cases}
\min(\rho A,\bar\rho A), & A\geq0,\\
\max\{\min(\rho A,\bar\rho A),CA\}, & A<0,
\end{cases}
\label{eq:supp-dual-clip-v3}
\end{equation}
The surrogate in Equation~\ref{eq:supp-dual-clip-v3} uses $\bar\rho=\operatorname{clip}(\rho,1-\epsilon,1+\epsilon)$. The reference-policy term is $d^{\mathrm{LV}}=e^\Delta-\Delta-1$, where $\Delta=\log\pi_{\mathrm{ref}}-\log\pi_\theta$, with numerical clipping. We use $\epsilon=0.2$, $C=3$, and $\beta=10^{-2}$ \citep{shao2024deepseekmath,schulman2017ppo,ye2020dualclip}.

The shared semantic reward is
\begin{equation}
r=0.6r_{\mathrm{acc}}+0.3(r_{\mathrm{logic}}+r_{\mathrm{prof}}+r_{\mathrm{concise}})/3+0.1r_{\mathrm{format}}.
\label{eq:supp-reward-v3}
\end{equation}
A locally deployed Qwen3.5-35B judge scores the semantic components, while normalized exact matches take a deterministic unit-score fast path. Table~\ref{tab:reward-components-v3} defines each component. The direct-answer template in the controlled experiment does not request the tags expected by the format regex. The selected endpoint logs record zero format reward for all three controlled arms, leaving a common accuracy-and-quality comparison.

\begin{table*}[t]
\centering
\small
\begin{tabular}{p{0.16\textwidth}p{0.09\textwidth}p{0.62\textwidth}}
\toprule
Component & Range & Operational meaning \\
\midrule
$r_{\mathrm{acc}}$ & $[0,1]$ & Factual correctness against the reference answer; normalized exact matches receive 1 \\
$r_{\mathrm{logic}}$ & $[0,1]$ & Logical soundness and consistency of the response \\
$r_{\mathrm{prof}}$ & $[0,1]$ & Professional, domain-appropriate expression \\
$r_{\mathrm{concise}}$ & $[0,1]$ & Absence of irrelevant or redundant content \\
$r_{\mathrm{format}}$ & $\{0,1\}$ & Presence of the answer tags required by the main-run response template \\
\bottomrule
\end{tabular}
\caption{Reward components used in Equations~\ref{eq:supp-advantage-v3} and~\ref{eq:supp-reward-v3}. The three semantic quality scores are averaged before receiving total weight 0.3.}
\label{tab:reward-components-v3}
\end{table*}

Each controlled arm runs updates 1--50 and then retains model weights while restarting the optimizer for updates 51--100. Residual-error sampling has the highest training-side validation accuracy reward at update 100 and is used for the main Wnuan-RL configuration. The main run and controlled experiment differ in response format, batch size, sequence length, hardware, and total schedule. Only the three controlled arms isolate data selection. The full-run online validation trajectory and stopping evidence are reported separately from the controlled comparison.

\subsection{Retrieval-Augmented Inference}

The RAG corpus contains 8,573 unique source files and 597,574 indexed chunks. Retrieval draws 30 candidates from BM25 \citep{robertson2009bm25} and 30 from BGE-M3 \citep{chen2024bgem3}, fuses them with weighted reciprocal-rank fusion, and retains five chunks. These corpus counts are distinct from the Stage-I training-data provenance counts.

Generation disables explicit thinking and uses temperature 0.7, top-$p$ 0.95, repetition penalty 1.1, at most 1,024 new tokens, and a 16,384-token maximum context. Retrieval traces are saved independently so that Base, Inst, and RL receive identical contexts. The no-RAG responses are historical generations rather than same-seed paired samples. No gold retrieval relevance labels or Recall@5 values are available.

\section{Stages I--II: Configuration Evidence}

\subsection{Stage-I Configurations}

\begin{table*}[t]
\centering
\small
\begin{tabular}{lrrrrrr}
\toprule
SFT data & AAR & Correct. & Complete. & Faithful. & Halluc. $\downarrow$ & General avg. \\
\midrule
Fixed-window text & 52.33 & 33.73 & 22.70 & 43.21 & 57.14 & 82.13 \\
Document-to-QA & \textbf{83.45} & \textbf{68.81} & \textbf{61.88} & \textbf{67.96} & \textbf{29.56} & 79.21 \\
Document-to-QA + answer rewriting & 82.04 & 64.71 & 56.01 & 63.44 & 32.11 & \textbf{82.15} \\
\bottomrule
\end{tabular}
\caption{End-to-end Stage-I configuration results (\%). Document-to-QA uses 27.8\% more estimated FLOPs than fixed-window training.}
\label{tab:sft-data-results-v3}
\end{table*}

\begin{table*}[t]
\centering
\small
\begin{tabular}{lrrr}
\toprule
SFT data & Updates & FLOPs & Runtime (s) \\
\midrule
Fixed-window & 3,974 & $5.29\!\times\!10^{14}$ & 10,765 \\
Document-to-QA & 5,082 & $6.76\!\times\!10^{14}$ & 26,764 \\
+ answer rewriting & 4,434 & $5.90\!\times\!10^{14}$ & 23,437 \\
\bottomrule
\end{tabular}
\caption{Training budgets for the Stage-I configurations. All runs use three epochs. FLOPs use a common per-update estimate, while wall-clock runtimes are logged separately.}
\label{tab:sft-compute-v3}
\end{table*}

Tables~\ref{tab:sft-data-results-v3} and~\ref{tab:sft-compute-v3} provide the complete outcome and budget breakdown underlying the Stage-I summary. The Document-to-QA gain over fixed-window training has a 95\% CI of 27.30--34.94 points (McNemar $p<0.001$). Answer rewriting changes AAR by $-1.41$ points (95\% CI: $-4.53$--1.56; $p=0.423$) and the general-benchmark average by +2.94 points. These are unequal-budget configuration studies: the logged runtimes differ more than the update and FLOP estimates.

\subsection{Stage-II Replay Grid}

\begin{table*}[t]
\centering
\small
\begin{tabular}{lrrrrrrr}
\toprule
Replay label & General examples & Actual ratio & Val. AAR & WnuanBench AAR & Correct. & Halluc. $\downarrow$ & General avg. \\
\midrule
0\% & 0 & 0.0\% & 76.78 & \textbf{82.04} & 64.71 & 32.11 & 82.15 \\
5\% & 10,000 & 4.5\% & 76.56 & 80.62 & 65.42 & 31.54 & 83.27 \\
25\% & 50,000 & 22.6\% & 76.56 & 81.05 & 63.93 & 31.97 & 83.65 \\
50\% & 106,950 & 48.3\% & \textbf{76.89} & 80.06 & 62.87 & 33.66 & \textbf{84.64} \\
100\% & 213,900 & 96.7\% & 76.44 & 80.34 & 63.51 & 31.12 & 83.31 \\
\bottomrule
\end{tabular}
\caption{General-data replay grid (\%). Replay labels are nominal display labels relative to 221,294 domain examples. The 48.3\% setting maximizes the public-general average and is selected before WnuanBench evaluation.}
\label{tab:replay-grid-v3}
\end{table*}

\begin{table*}[t]
\centering
\small
\begin{tabular}{lrrr}
\toprule
Actual replay ratio & MMLU & IFEval & C-Eval \\
\midrule
0.0\% & 85.58 & 79.00 & 81.87 \\
4.5\% & 85.02 & 83.15 & 81.65 \\
22.6\% & 86.32 & 82.02 & 82.62 \\
48.3\% & 85.53 & \textbf{86.52} & 81.88 \\
96.7\% & \textbf{86.65} & 80.00 & \textbf{83.28} \\
\bottomrule
\end{tabular}
\caption{Components of the general-benchmark average in Table~\ref{tab:replay-grid-v3}.}
\label{tab:replay-components-v3}
\end{table*}

Tables~\ref{tab:replay-grid-v3} and~\ref{tab:replay-components-v3} give the complete replay grid behind the selected 48.3\% operating point. Selection uses the public-general average, with the validation result as a secondary development measure. The validation and later WnuanBench rankings differ across the five candidates (Pearson $r=0.12$, Spearman $\rho=-0.05$). For budget context, a six-epoch domain-only reference obtains 85.86\% AAR and an 80.37\% general average with 8,868 updates and $1.18\times10^{15}$ FLOPs. The 96.7\% replay run obtains 80.34\% and 83.31\% with 11,976 updates and $1.59\times10^{15}$ FLOPs, so the 35\% FLOP difference precludes a compute-matched interpretation.

\section{Stage III: Controlled Residual-Error Sampling}

\subsection{Data Arms and Endpoints}

Residual selection uses a 230,183-row QA pool with original targets. The recorded adaptive three-model protocol labels 56,147 Wnuan-Inst responses incorrect after conditional disagreement adjudication and ordered-median aggregation. The residual arm uses those rows. The full arm samples from all 230,183 rows, and the random arm uses seed 42 to draw 56,147 rows from the same pool. A fourth reward-sensitivity arm keeps the residual pool but changes the accuracy/quality weights from 0.6/0.3 to 0.7/0.2. It is not a data-selection control.

\begin{table*}[t]
\centering
\small
\begin{tabular}{lrrrrrr}
\toprule
Arm & Pool size & AAR & Correct. & Complete. & Faithful. & Halluc. $\downarrow$ \\
\midrule
Residual errors & 56,147 & \textbf{89.39} & \textbf{74.05} & 63.37 & \textbf{71.71} & 20.37 \\
Full pool & 230,183 & 86.28 & 72.14 & 64.00 & 68.95 & 24.61 \\
Size-matched random & 56,147 & 86.42 & 72.56 & \textbf{66.55} & 68.10 & 23.20 \\
Reward 0.7/0.2 & 56,147 & 89.25 & 74.12 & 62.38 & 71.29 & \textbf{20.08} \\
\bottomrule
\end{tabular}
\caption{Controlled answer-only GRPO endpoints at update 100 (\%).}
\label{tab:rl-results-v3}
\end{table*}

Table~\ref{tab:rl-results-v3} reports the update-100 WnuanBench endpoints. After fixing those checkpoints and the evaluation recipe, we evaluate the same endpoints on the same-domain validation set. Generation settings and formal scoring are common across arms: gpt-oss-120b and MiniMax-M2.5 provide the primary correctness votes, DeepSeek-V3.2 adjudicates disagreements, and gpt-oss-120b scores the auxiliary dimensions. Tables~\ref{tab:rl-validation-results-v3} and~\ref{tab:rl-validation-paired-v3} report this retrospective comparison. It tests consistency within the same enterprise distribution, not source-held-out evidence.

\begin{table*}[t]
\centering
\small
\begin{tabular}{lrrrrrr}
\toprule
Arm & AAR & 95\% CI & Correct. & Complete. & Faithful. & Halluc. $\downarrow$ \\
\midrule
Residual errors & \textbf{81.33} & 78.78--83.78 & \textbf{63.89} & 44.56 & \textbf{41.39} & \textbf{24.67} \\
Full pool & 77.78 & 75.11--80.56 & 62.67 & \textbf{49.00} & 37.78 & 31.56 \\
Size-matched random & 78.67 & 76.00--81.33 & 62.83 & 48.89 & 37.44 & 33.00 \\
\bottomrule
\end{tabular}
\caption{Retrospective Stage-III endpoints on the validation set (\%). Intervals resample questions from each fixed endpoint and do not capture retraining variance.}
\label{tab:rl-validation-results-v3}
\end{table*}

\begin{table*}[t]
\centering
\small
\begin{tabular}{lrrrrr}
\toprule
Comparison & Difference & 95\% CI & Discordant pairs & Raw $p$ & Holm $p$ \\
\midrule
Residual $-$ full & +3.56 & +1.33 to +5.78 & 69 / 37 & 0.0026 & 0.0078 \\
Residual $-$ random & +2.67 & +0.22 to +5.00 & 71 / 47 & 0.0342 & 0.0685 \\
Full $-$ random & $-0.89$ & $-3.22$ to +1.33 & 53 / 61 & 0.5121 & 0.5121 \\
\bottomrule
\end{tabular}
\caption{Paired validation-set comparisons. Differences are AAR percentage points. Discordant pairs list improvements/regressions for the first-named arm. Holm adjustment covers all three contrasts.}
\label{tab:rl-validation-paired-v3}
\end{table*}

\subsection{Reference-Target Sensitivity Diagnostic}
\label{sec:residual-target-sensitivity-v3}

Residual selection uses the original targets in the 230,183-row selection pool. To assess whether automated incorrect labels are sensitive to that reference choice, we use an archived rewrite-output artifact, prior to final exact-question exclusion, to map a unique alternative target to 231,512 of the 231,662 inventory rows. Of these comparable rows, 107,432 have a changed target and 124,080 are identical exactly or after normalization; 150 rows without a unique alternative target are excluded. We hold each archived Wnuan-Inst prediction fixed and re-judge all 107,432 changed-target rows against the rewritten target using the formal three-judge correctness ensemble. The rescan contains 107,432 valid results, with no missing rows, judge errors, or invalid or duplicate sample identifiers.

\begin{table*}[t]
\centering
\small
\begin{tabular}{lrr}
\toprule
Statistic & Changed targets & All comparable targets \\
\midrule
Rows & 107,432 & 231,512 \\
Incorrect under both targets & 22,086 & 51,207 \\
Original-only incorrect & 4,928 & 4,928 \\
Rewritten-only incorrect & 4,001 & 4,001 \\
Neither incorrect & 76,417 & 171,376 \\
Membership unchanged (\%) & 91.69 & 96.14 \\
Original-target incorrect rate (\%) & 25.15 & 24.25 \\
Rewritten-target incorrect rate (\%) & 24.28 & 23.85 \\
Incorrect-set Jaccard overlap & 0.712 & 0.852 \\
\bottomrule
\end{tabular}
\caption{Fixed-prediction reference-target sensitivity. The changed-target column is the primary diagnostic. The all-comparable column additionally reuses the original label for 124,080 unchanged targets and therefore mechanically has higher agreement. Rows without a unique rewritten target ($n=150$) are excluded.}
\label{tab:residual-target-sensitivity-v3}
\end{table*}

Table~\ref{tab:residual-target-sensitivity-v3} shows that changed-target membership is stable for 91.69\% of rows and that the incorrect rate changes by $-0.86$ percentage points, from 25.15\% to 24.28\%. The nonzero migration is bidirectional: 4,928 rows leave and 4,001 enter the automated incorrect set. Across all comparable rows, membership agreement is 96.14\%, but this aggregate includes the 124,080 rows whose targets did not change. As a separate provenance check on the pre-exclusion archive, accepted-rewrite rows have a 23.75\% historical residual rate, compared with 25.77\% for retained-target rows; this association does not support systematic over-selection of accepted rewrites and is not interpreted causally.

This fixed-prediction audit measures reference-target sensitivity under the shared three-judge aggregation policy. Re-judging only the rewritten-target side leaves judge rerun variability in the observed migrations, so the audit cannot isolate a causal effect of rewriting. It also does not replace human semantic-equivalence validation or a comparison of models trained on original and rewritten targets.

\subsection{Source-Cluster Bootstrap Sensitivity}
\label{sec:source-cluster-bootstrap-v3}

\begin{table*}[t]
\centering
\small
\begin{tabular}{lrrr}
\toprule
Paired contrast & Difference & Question-level 95\% CI & Source-cluster 95\% CI \\
\midrule
Wnuan-Inst $-$ Wnuan-Base & +27.30 & +23.20 to +31.54 & +19.71 to +34.21 \\
Wnuan-RL $-$ Wnuan-Inst & +11.45 & +8.49 to +14.43 & +8.52 to +15.05 \\
Residual $-$ full & +3.11 & +0.71 to +5.66 & +0.83 to +5.70 \\
Residual $-$ random & +2.97 & +0.85 to +5.09 & +0.81 to +5.08 \\
Full $-$ random & $-0.14$ & $-2.55$ to +2.12 & $-2.64$ to +2.04 \\
\bottomrule
\end{tabular}
\caption{Question- and source-cluster bootstrap sensitivity on WnuanBench (AAR percentage points). Primary question-level intervals use 2,000 resamples. Source-cluster intervals use 10,000 resamples over 217 source documents, retaining all questions from each sampled document.}
\label{tab:source-cluster-bootstrap-v3}
\end{table*}

Table~\ref{tab:source-cluster-bootstrap-v3} accounts for correlation among questions grounded in the same source document. Both pipeline-stage gains and both residual-versus-control contrasts remain above zero under source-cluster resampling. The full-versus-random interval spans zero under both resampling schemes.

Under the prespecified ensemble, the complete arm ordering matches WnuanBench: residual first, random second, and full third. Across only three fixed arms, Pearson $r=0.98$ and Spearman $\rho=1.00$ are descriptive consistency checks, not population-level correlation evidence. Residual selection ranks first in five of nine validation domains and five of eight WnuanBench domains. The taxonomies differ, so we do not align domains across sets. The residual--full contrast survives multiplicity correction on both sets. The residual--random validation interval excludes zero before correction, but its Holm-adjusted $p=0.0685$ does not. We treat this result as directionally consistent rather than a second significant replication.

\subsection{Aggregate GRPO Signal Diagnostics}

The W\&B histories contain update-level scalar aggregates for the main run and all three controlled arms, without prompt or response text. Table~\ref{tab:grpo-reward-decomposition-v3} applies Equation~\ref{eq:supp-reward-v3} to the main-run validation changes from update 0 to update 327. The accuracy, mean-quality, and format components change by 0.1964, 0.1384, and 0.7866. After weighting, they contribute 0.1179, 0.0415, and 0.0787 to the 0.2380 overall-reward gain.

\begin{table*}[t]
\centering
\small
\begin{tabular}{lrrrrr}
\toprule
Reward component & Coefficient & Update 0 & Update 327 & Raw change & Weighted contribution \\
\midrule
Accuracy & 0.6 & 0.5836 & 0.7800 & +0.1964 & +0.1179 \\
Mean of logic, professionalism, and conciseness & 0.3 & 0.7457 & 0.8840 & +0.1384 & +0.0415 \\
Format & 0.1 & 0.2134 & 1.0000 & +0.7866 & +0.0787 \\
\midrule
Overall reward & -- & 0.5952 & 0.8332 & +0.2380 & +0.2380 \\
\bottomrule
\end{tabular}
\caption{Main-run validation-reward decomposition from update 0 to update 327. Weighted contribution is the coefficient multiplied by the raw component change. The three component contributions sum to the observed overall-reward change.}
\label{tab:grpo-reward-decomposition-v3}
\end{table*}

\begin{figure*}[!t]
\centering
\includegraphics[width=\textwidth]{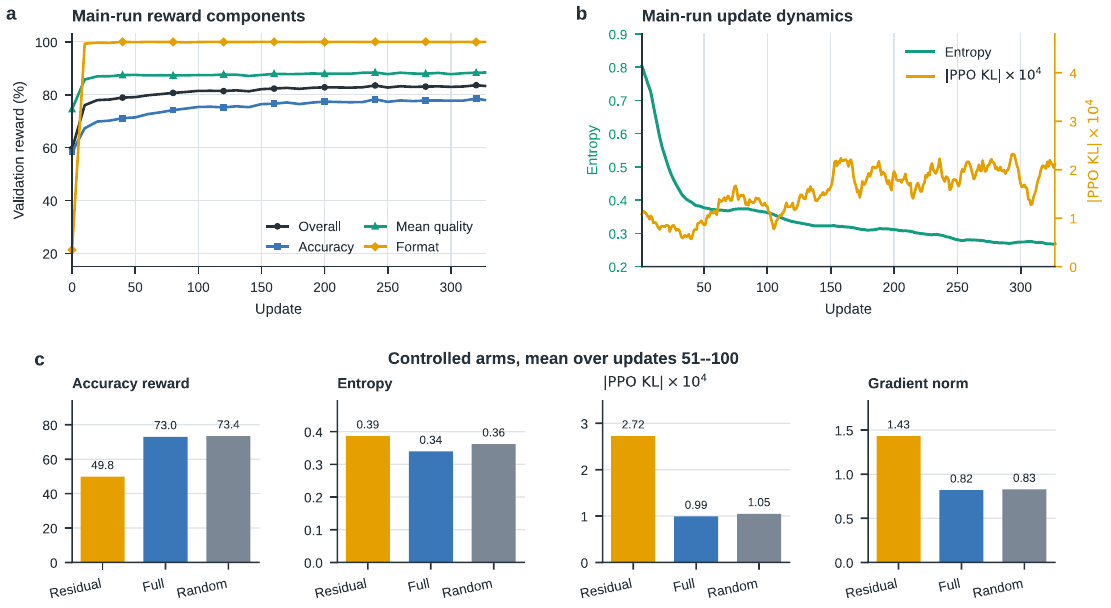}
\caption{Aggregate GRPO diagnostics. (a) Main-run validation reward components. (b) Main-run entropy and absolute PPO-KL, shown as centered 15-update moving averages for readability. (c) Raw controlled-arm means over updates 51--100. The residual arm combines lower on-policy accuracy reward with larger aggregate update statistics under the matched protocol.}
\label{fig:grpo-signal-dynamics-v3}
\end{figure*}

Figure~\ref{fig:grpo-signal-dynamics-v3} summarizes the main-run and controlled-arm signals. Under the matched protocol, the residual arm receives lower mean on-policy accuracy reward over updates 51--100 (0.498 versus 0.730 for full-pool and 0.734 for random sampling), indicating harder sampled prompts. It also records higher entropy, mean absolute PPO-KL, gradient norm, and upper-clipping fraction (Table~\ref{tab:grpo-controlled-signals-v3}). These update-level statistics describe training dynamics, not causal mediation.

\begin{table*}[t]
\centering
\small
\begin{tabular}{lrrrrr}
\toprule
Training arm & Accuracy reward & Entropy & $|\mathrm{PPO\ KL}|\times10^4$ & Gradient norm & Upper clip (\%) \\
\midrule
Residual errors & 0.4983 & 0.3863 & 2.724 & 1.432 & 0.958 \\
Full pool & 0.7299 & 0.3394 & 0.992 & 0.819 & 0.718 \\
Size-matched random & 0.7341 & 0.3619 & 1.049 & 0.827 & 0.721 \\
\bottomrule
\end{tabular}
\caption{Mean W\&B scalars over controlled updates 51--100. The three runs match on seed, rollouts per prompt, batch sizes, learning rate, PPO epochs, response limit, validation frequency, GRPO estimator, and KL coefficient.}
\label{tab:grpo-controlled-signals-v3}
\end{table*}

\begin{figure*}[!t]
\centering
\includegraphics[width=\textwidth]{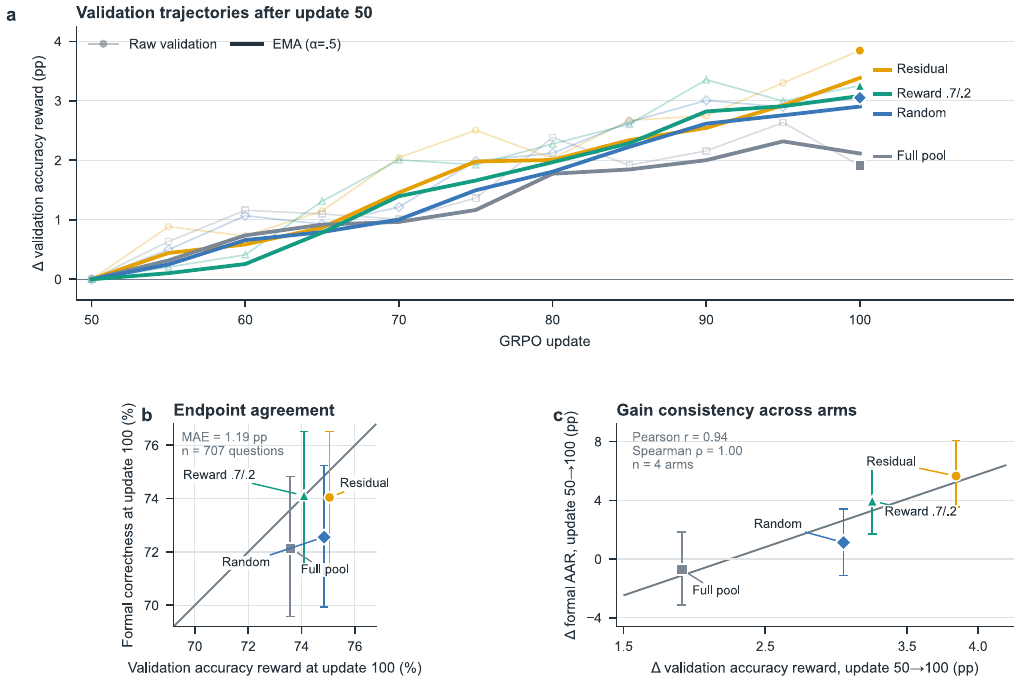}
\caption{Post hoc agreement between the Stage-III training monitor and final evaluation. (a) Change in the raw validation accuracy reward from update 50, with an EMA overlay for readability. (b) Raw update-100 validation accuracy reward versus final formal correctness on the same 707 WnuanBench questions. The dashed line denotes equality, and error bars are 95\% question-level bootstrap intervals from 2,000 resamples. The mean absolute discrepancy is 1.19 percentage points. (c) Update-50-to-100 change in the validation signal versus the paired change in formal AAR. Error bars are paired 95\% question-level bootstrap intervals. Pearson $r=0.94$ and Spearman $\rho=1.00$ are descriptive across four fixed arms, not estimates of retraining variability.}
\label{fig:supp-stage3-training-monitor}
\end{figure*}

Figure~\ref{fig:supp-stage3-training-monitor} compares the online validation trace with the final WnuanBench endpoint. The validation signal is logged every five updates after update 50. Panel (a) subtracts each arm's update-50 value, and the exponential moving average ($\alpha=0.5$) is used only for visualization. All calculations use unsmoothed records.

At update 50, the residual, full, and random arms obtain 83.73\%, 86.99\%, and 85.29\% AAR. By update 100, their AAR changes by +5.66, $-0.71$, and +1.13 points, respectively. These changes cover the common second segment after the optimizer restart.

The online monitor is a continuous reward-judge average, whereas formal correctness is ordinal and AAR thresholds the final ensemble label. Their mean endpoint discrepancy is 1.19 points at update 100, and relative monitor changes preserve the observed ordering of AAR gains. The four completed arms are insufficient to estimate a general correlation or reconstruct a pointwise AAR training curve.

\begin{table*}[t]
\centering
\small
\begin{tabular}{lrrrrr}
\toprule
Comparison & Difference & 95\% CI & Discordant pairs & Raw $p$ & Holm $p$ \\
\midrule
Residual $-$ full & +3.11 & +0.71 to +5.66 & 50 / 28 & 0.017 & 0.035 \\
Residual $-$ random & +2.97 & +0.85 to +5.09 & 42 / 21 & 0.012 & 0.035 \\
Full $-$ random & $-0.14$ & $-2.55$ to +2.12 & 35 / 36 & 1.000 & 1.000 \\
\bottomrule
\end{tabular}
\caption{Planned paired comparisons at update 100. Differences and intervals are percentage points. Discordant pairs list improvements/regressions for the first-named arm.}
\label{tab:rl-paired-v3}
\end{table*}

Table~\ref{tab:rl-paired-v3} gives the planned WnuanBench contrasts. The update-100 AAR intervals are 87.1--91.5 for residual, 83.7--88.8 for full, and 84.0--88.8 for random. Residual versus random controls pool size, while residual versus full holds updates fixed but not per-example exposure. Full and random are statistically indistinguishable on both question sets. The 0.14-point gap between the default and 0.7/0.2 reward arms does not establish robustness to reward weights.

\section{Capability Retention and Residual Errors}

\begin{figure*}[!t]
\centering
\includegraphics[width=\textwidth]{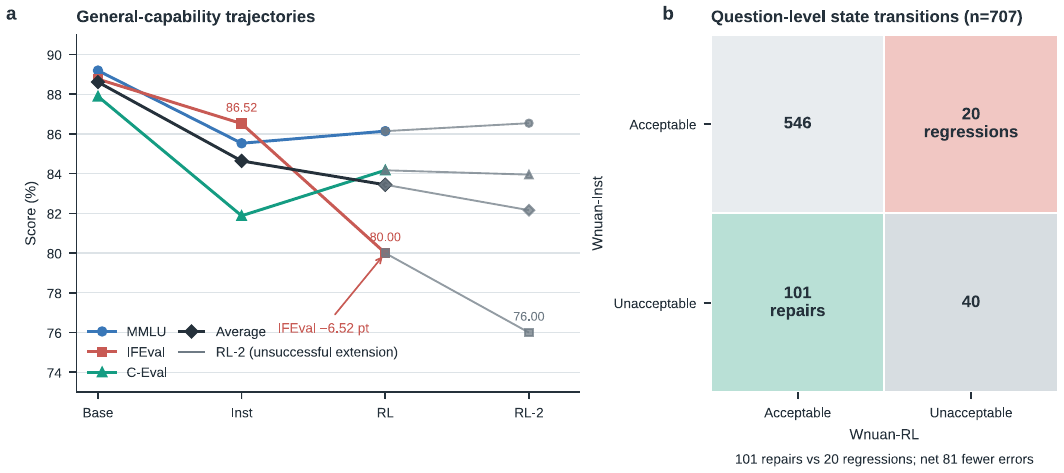}
\caption{Capability retention and question-level transitions. (a) MMLU, IFEval, C-Eval, and their unweighted average across Base, Inst, and RL. The unsuccessful RL-2 extension is de-emphasized in gray. The Inst-to-RL average changes by $-1.20$ points, while IFEval changes by $-6.52$ points. (b) Wnuan-Inst-to-Wnuan-RL transitions on the same 707 questions: 101 repairs, 20 regressions, 546 retained acceptable answers, and 40 persistent failures.}
\label{fig:supp-capability-transition-v3}
\end{figure*}

\begin{table*}[t]
\centering
\small
\begin{tabular}{llrrrr}
\toprule
Route & Checkpoint & MMLU & IFEval & C-Eval & General avg. \\
\midrule
32B & Wnuan-Base & 89.19 & 88.76 & 87.89 & 88.61 \\
32B & Wnuan-Inst & 85.53 & 86.52 & 81.88 & 84.64 \\
32B & Wnuan-RL & 86.14 & 80.00 & 84.17 & 83.44 \\
\midrule
671B & Wnuan-Plus-Base & 91.82 & 88.00 & 91.31 & 90.38 \\
671B & Wnuan-Plus-Inst & 86.53 & 83.00 & 83.49 & 84.34 \\
\bottomrule
\end{tabular}
\caption{Public-benchmark components for the two Wnuan routes (\%). General avg. is the unweighted mean of MMLU, IFEval, and C-Eval. Cross-route values are descriptive because the routes use different backbones and adaptation procedures.}
\label{tab:general-components-v3}
\end{table*}

Figure~\ref{fig:supp-capability-transition-v3}(a) provides the full 32B public-benchmark trajectory behind the aggregate retention result, and Table~\ref{tab:general-components-v3} reports the endpoint components for both routes. Within the 32B route, the Inst-to-RL average decline is concentrated in IFEval rather than shared uniformly across MMLU, IFEval, and C-Eval. Within the separate Wnuan-Plus route, all three components decrease after LoRA-SFT: MMLU by 5.29 points, IFEval by 5.00 points, and C-Eval by 7.82 points. Panel (b) gives the complete acceptable/unacceptable transition accounting for Wnuan-Inst to Wnuan-RL. Qualitative review localizes recurrent failures to atomic numbers and dates, departmental ownership, exact document names, closed-set enumerations, and conditions that delimit otherwise correct rules. Judge rationales also suggest factual substitution, missing required points, and unsupported answer expansion. These categories are not reported as frequencies because WnuanBench lacks mutually exclusive expert error labels.

\section{Within-Enterprise Domain Analysis}

\begin{figure*}[!t]
\centering
\includegraphics[width=0.82\textwidth]{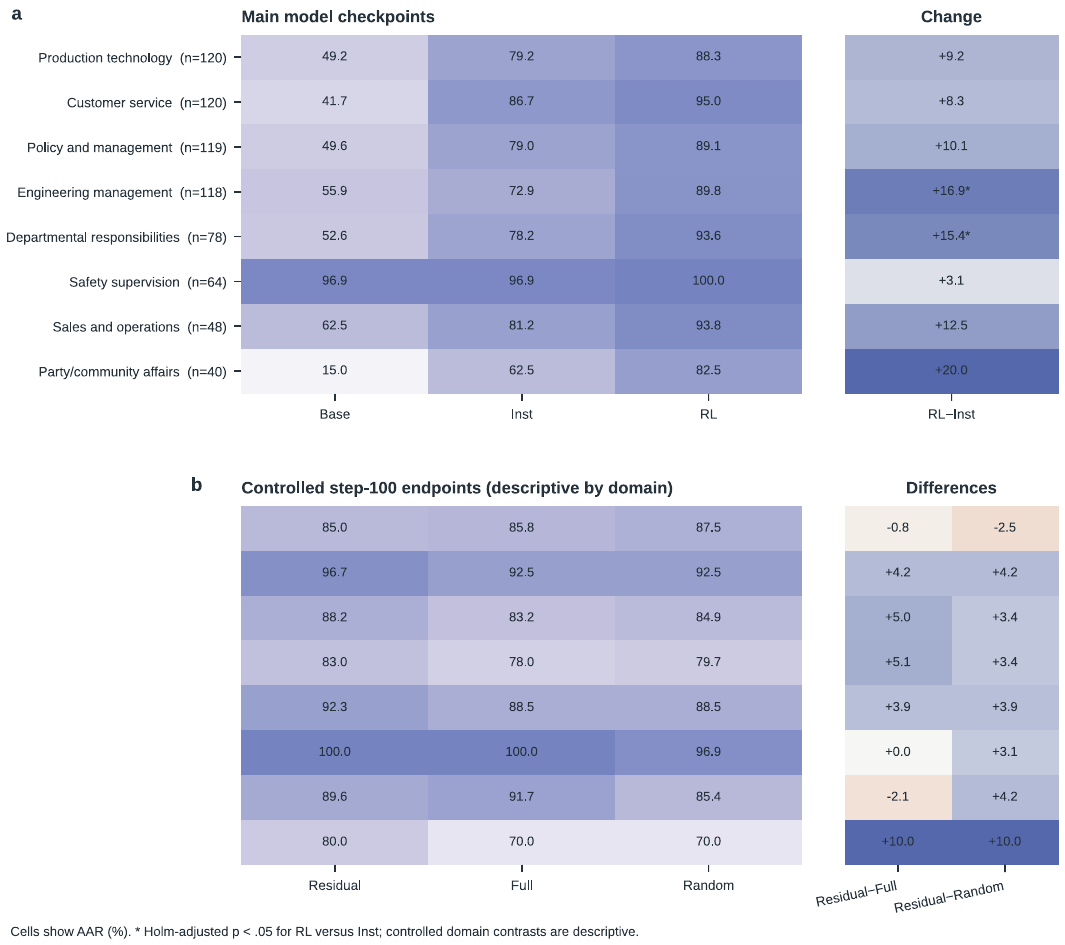}
\caption{Business-domain results. (a) Base, Inst, and RL AAR and the RL-minus-Inst change. All eight changes are positive. Asterisks mark the two domains with Holm-adjusted $p<0.05$. (b) Controlled residual, full, and random endpoints with residual-minus-control differences. No controlled domain contrast is significant, so these slices are descriptive.}
\label{fig:supp-domain-v3}
\end{figure*}

Figure~\ref{fig:supp-domain-v3} provides the complete descriptive domain breakdown. Six raw checkpoint-trajectory McNemar tests are below 0.05, but only engineering management and departmental responsibilities remain significant after Holm correction. In the controlled experiment, production technology and sales are the two numerical exceptions relative to full-pool sampling, and none of the domain-level controlled contrasts is significant. Small slices and ceiling effects make the aggregate comparisons primary.

\section{Extended RAG Results}

\begin{figure*}[!t]
\centering
\includegraphics[width=0.83\textwidth]{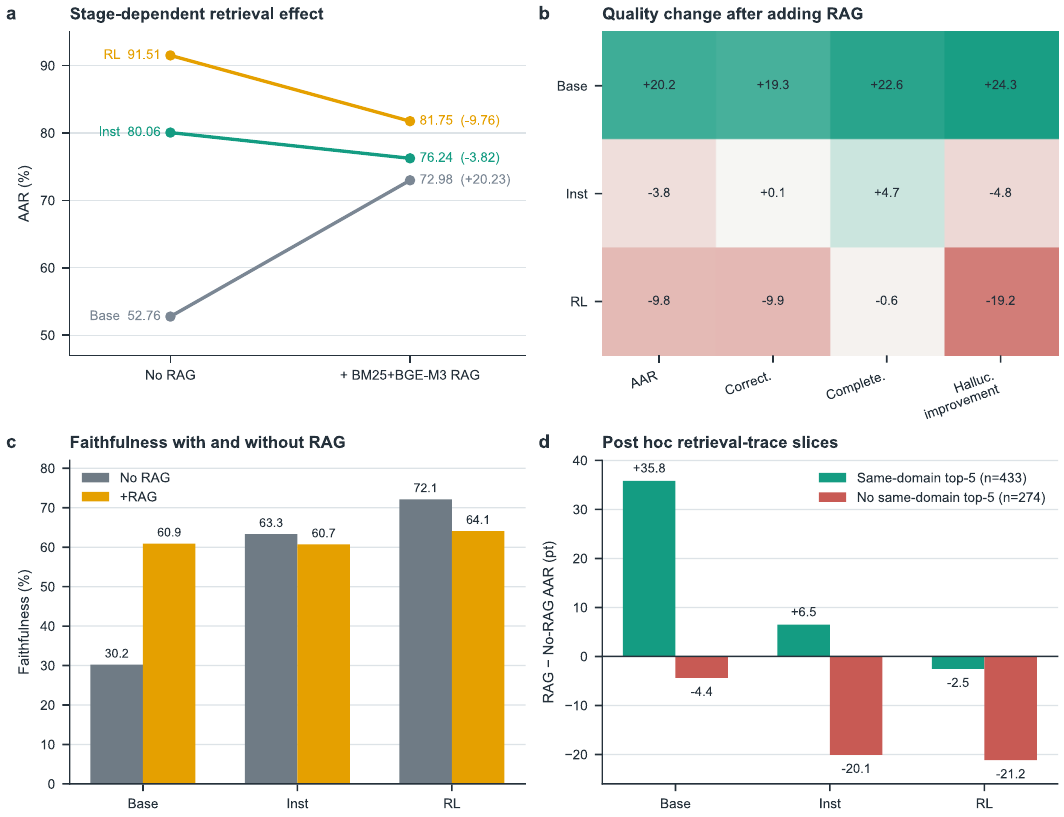}
\caption{RAG diagnostics. (a) AAR before and after applying the saved BM25+BGE-M3 top-5 traces. (b) Signed changes in AAR, correctness, completeness, and hallucination. Hallucination is sign-inverted so positive means improvement. (c) Stored no-RAG and reported RAG faithfulness under the common aggregation protocol. (d) RAG-minus-no-RAG AAR in post hoc retrieval-trace slices. Same-domain matching is a proxy, not gold Recall@5.}
\label{fig:supp-rag-v3}
\end{figure*}

\subsection{Same-Question Comparisons}

RAG changes Wnuan-Base from 52.76\% to 72.98\% AAR, Wnuan-Inst from 80.06\% to 76.24\%, and Wnuan-RL from 91.51\% to 81.75\%. Improved/regressed item counts are 198/55, 76/103, and 29/98. The exact McNemar $p$-values are $<0.0001$, 0.0517, and $<0.0001$. Under the common retrieval traces, the SFT and RL stage gains remain +3.25 and +5.52 points, smaller than their no-RAG counterparts.

The no-RAG generations were not regenerated with the same decoding seeds. The results show a stage-dependent end-to-end pattern under one retrieval system, not a randomized retrieval-by-training interaction.

\subsection{Quality Dimensions}

Figure~\ref{fig:supp-rag-v3}(b) reports the changes in correctness, completeness, and hallucination under RAG. Panel (c) compares the stored no-RAG and reported RAG faithfulness aggregates under the common evaluation protocol.

\subsection{Post Hoc Retrieval-Trace Slices}

At least one retrieved chunk shares the question's business domain for 433 questions. No top-five chunk shares the domain for 274. In the same-domain slice, RAG changes Base, Inst, and RL by +35.80, +6.47, and $-2.54$ points. In the proxy-negative slice, the changes are $-4.38$, $-20.07$, and $-21.17$ points. A domain match need not contain the answer and may correlate with question difficulty. The association motivates retrieval confidence gating but does not establish retrieval mismatch as the cause.

\section{API Context and Wnuan-Plus Configuration}

Table 1 of the main paper reports six API endpoints and the Wnuan-Plus route. The API systems are identified by their official provider releases \citep{zai2026glm51,minimax2026m3,moonshotai2026kimik26,deepseekai2026v4,xiaomi2026mimov2pro,openai2026gpt54}. Their decoding, serving configuration, and compute are not matched, so they provide same-question context rather than a controlled model comparison. Wnuan-Plus starts from DeepSeek-V3.1-Terminus \citep{deepseekai2025v31terminus}, receives LoRA-SFT but no RL, and records the SFT route at a distinct scale. It does not isolate model size, data, or adaptation method. Supplementary Appendix G reports the corresponding general-capability components.

\begin{table*}[t]
\centering
\small
\begin{tabular}{p{0.22\textwidth}p{0.69\textwidth}}
\toprule
Component & Recorded configuration \\
\midrule
Model and adaptation & Wnuan-Plus-Inst; initialized from DeepSeek-V3.1-Terminus; LoRA-SFT only; no RL \\
Budget and precision & 3 epochs; 30,540 updates; bf16; maximum sequence length 700 \\
Batching & Global batch 3; microbatch 1; gradient accumulation 1 \\
Optimization & HybridAdam; peak LR $10^{-5}$; warmup 0.05; weight decay 0.1; gradient clipping at 1.0 \\
LoRA & Rank 16; alpha 32 \\
Parallelism & 3 nodes $\times$ 8 accelerators; tensor parallel 1; pipeline parallel 3; expert parallel 8 \\
Memory and kernels & ZeRO-2 with CPU offload; gradient checkpointing; FlashAttention \\
\bottomrule
\end{tabular}
\caption{Recorded configuration of the 671B Wnuan-Plus route. The available run metadata do not identify the accelerator model.}
\label{tab:671b-config-v3}
\end{table*}

Table~\ref{tab:671b-config-v3} records the available Wnuan-Plus configuration.

\section{Unsuccessful Extension Beyond Stage III}

Wnuan-RL-2 starts from Wnuan-RL and partitions the 231,662 pre-rewriting-inventory examples according to Wnuan-Inst and Wnuan-RL correctness: persistent errors (A), regressions (B), learned cases (C), and consistently correct cases (D). It retains all 19,401 A and 9,951 B examples, 3,675 of 36,746 C examples (10\%), and 4,967 of 165,564 D examples (3\%), for 37,994 examples in total.

The endpoint changes AAR from 91.51\% to 91.37\%, correctness from 78.43\% to 79.70\%, completeness from 66.76\% to 69.73\%, faithfulness from 72.14\% to 69.66\%, hallucination from 15.70\% to 20.93\%, and IFEval from 80.00\% to 76.00\%. AAR is statistically unchanged ($-0.14$ points, McNemar $p=1$), while hallucination and instruction following worsen. Without an otherwise identical unbucketed control, this experiment does not isolate the partition rule. It shows only that the evaluated continuation is not a successful extension.

\section{Reproducibility and Responsible Use}

\subsection{Releaseable Evaluation Capsule}

The separately uploaded Code and Data Supplement includes a non-proprietary evaluation capsule under \path{reproducibility/}. The file \path{correctness_judge_prompt.txt} gives the complete correctness prompt used for the primary outcome, including the $0/0.5/1$ rubric and JSON output contract. The definitions for completeness, faithfulness, and hallucination appear in Supplementary Appendix A. These are supporting dimensions rather than the primary inference.

The evaluation flow is:
\begin{enumerate}
\item generate one answer per frozen question and checkpoint under the recorded decoding condition;
\item score correctness independently with gpt-oss-120b and MiniMax-M2.5 using the released template;
\item if the two ordered labels disagree, obtain a DeepSeek-V3.2 vote and retain the median of the three labels;
\item map scores $1$, $0.5$, and $0$ to correct, partially correct, and incorrect, and compute AAR by the definition in the main paper;
\item compare aligned questions with paired bootstrap intervals and McNemar tests, applying Holm correction within the planned three-arm family.
\end{enumerate}

\path{synthetic_evaluation_fixture.jsonl} provides four explicitly fictional question--reference--prediction pairs with paired endpoint labels. It contains no enterprise document, question, answer, evidence, path, or identifier. The standard-library script \path{reference_statistics.py} validates the schema and reproduces AAR, mean ordered correctness, a 2,000-resample paired bootstrap interval with seed 20260708, and an exact McNemar test. The bundled script \path{reproducibility/analyze_stage3_validation.py} regenerates the reported cross-set endpoint tables. \path{reproducibility/analyze_source_cluster_bootstrap.py} regenerates the question- and source-cluster sensitivity table, while \path{reproducibility/analyze_training_benchmark_overlap.py} removes exact question matches before computing text-free, hash-indexed BGE-M3 nearest-neighbor statistics. With authorized read access to the recorded runs, \path{reproducibility/analyze_wandb_grpo_signals.py} requests scalar keys only and regenerates the aggregate GRPO trajectories, reward decomposition, controlled-arm summary, and configuration summary. Their private inputs are described by the command-line interfaces and are not included in the release.

\subsection{Data Availability}

The private source documents and complete WnuanBench cannot be distributed because they are governed enterprise materials. The released materials expose the primary evaluation prompt, label aggregation, result schema, statistical transformations, training configurations, data roles, endpoint counts, paired tests, and snapshot digest prefixes. Row-level enterprise provenance and several historical environment fields remain inside the controlled archive.

\subsection{Local Processing and Data Governance}

The study uses authorized internal policy, standard, and process documents under data-minimization and de-identification procedures. Document processing, QA construction, filtering, target rewriting, SFT, residual selection, and GRPO ran on locally deployed models inside the controlled environment. No external provider API was used for data construction or training, and no source document or evidence excerpt was transmitted outside that environment. The six API systems in Supplementary Appendix J received benchmark question text only, without source documents, evidence excerpts, or reference answers.

\subsection{Intended Use}

Wnuan is intended as an internal knowledge assistant whose outputs require human verification. It is not intended to make automated personnel, compliance, safety, or other high-impact decisions. The evidence is limited to one enterprise, one same-domain validation set, one final in-domain benchmark, one completed run per configuration, and the reported retrieval and optimization budgets.

\end{document}